\documentclass{article} 
\usepackage{iclr2027_conference,times}

\usepackage{amsmath,amsfonts,bm}

\def\eqref#1{equation~\ref{#1}}

\def\1{\bm{1}}

\def\vzero{{\bm{0}}}

\def\vepsilon{{\bm{\epsilon}}}

\def\va{{\bm{a}}}

\def\vc{{\bm{c}}}

\def\vo{{\bm{o}}}

\def\vq{{\bm{q}}}

\def\vu{{\bm{u}}}
\def\vv{{\bm{v}}}

\def\vz{{\bm{z}}}

\def\mA{{\bm{A}}}

\def\mI{{\bm{I}}}

\def\mZ{{\bm{Z}}}

\DeclareMathAlphabet{\mathsfit}{\encodingdefault}{\sfdefault}{m}{sl}
\SetMathAlphabet{\mathsfit}{bold}{\encodingdefault}{\sfdefault}{bx}{n}

\def\gC{{\mathcal{C}}}

\def\gJ{{\mathcal{J}}}

\def\gL{{\mathcal{L}}}

\newcommand{\enc}{\operatorname{Enc}}
\newcommand{\statebranch}{f_z}
\newcommand{\actionbranch}{f_a}

\DeclareMathOperator*{\argmin}{arg\,min}

\usepackage[hidelinks,hypertexnames=false]{hyperref}
\usepackage{url,xspace}
\usepackage{graphicx}
\usepackage{booktabs}
\usepackage{multirow}
\usepackage{array}
\usepackage{colortbl}
\usepackage{pifont}
\usepackage{fontawesome5}
\usepackage{enumitem}
\usepackage{placeins}
\usepackage{float}
\usepackage{wrapfig}
\usepackage{needspace}
\usepackage{algorithm}
\usepackage{algpseudocode}
\usepackage{tikz}
\usetikzlibrary{arrows.meta,positioning,fit,calc,backgrounds}
\usepackage[most]{tcolorbox}

\setlist[itemize,1]{
  label=\textbullet,
  labelindent=0pt,
  leftmargin=*,
  labelsep=0.55em,
  topsep=0.35em,
  itemsep=0.25em,
  parsep=0pt,
  partopsep=0pt
}

\setlist[enumerate,1]{
  label=\arabic*.,
  labelindent=0pt,
  leftmargin=*,
  labelsep=0.55em,
  topsep=0.35em,
  itemsep=0.25em,
  parsep=0pt,
  partopsep=0pt
}

\newcommand{\finishwrapfigure}{%
  \par
  \ifnum\value{WF@wrappedlines}>1
    \vspace{\dimexpr\value{WF@wrappedlines}\baselineskip-\baselineskip\relax}%
  \fi
  \WFclear
}

\definecolor{PaperBlue}{HTML}{486B89}
\definecolor{PaperBlueFill}{HTML}{CBD9E3}
\definecolor{PaperGray}{HTML}{92989E}
\definecolor{PaperGrayFill}{HTML}{E2E5E7}
\definecolor{PaperGrayDark}{HTML}{68717A}
\definecolor{PaperGrayWash}{HTML}{F6F7F8}
\colorlet{LeWAMState}{PaperBlue}
\colorlet{LeWAMStateLight}{PaperBlueFill}
\definecolor{LeWAMStateEdge}{HTML}{93AABD}
\definecolor{LeWAMStateWash}{HTML}{F4F7FA}
\definecolor{LeWAMStatePale}{HTML}{F9FBFC}
\colorlet{LeWAMAction}{PaperGray}
\colorlet{LeWAMActionDark}{PaperGrayDark}
\colorlet{LeWAMActionLight}{PaperGrayFill}
\colorlet{LeWAMActionWash}{PaperGrayWash}
\colorlet{LeWAMObjective}{PaperGrayDark}
\definecolor{FigureInk}{HTML}{202124}
\definecolor{FigureMuted}{HTML}{6F747A}
\definecolor{FigureVisible}{HTML}{E9ECEF}
\definecolor{FigureGrid}{HTML}{A8ADB1}
\definecolor{FigureBorder}{HTML}{767C82}
\definecolor{FigureWhite}{HTML}{FFFFFF}
\definecolor{FigureModuleFill}{HTML}{FAFBFC}
\colorlet{FigureObjectiveFill}{PaperGrayWash}
\definecolor{FigureDivider}{HTML}{C5C7CA}
\definecolor{SurfaceNeutral}{HTML}{F7F6F3}
\definecolor{SurfaceEmphasis}{HTML}{EFF5F1}
\definecolor{SurfacePrior}{HTML}{F3EBDD}
\definecolor{SurfaceContribution}{HTML}{E1EEE6}
\definecolor{SurfaceRule}{HTML}{CDD3CF}
\definecolor{DiagramEncoder}{HTML}{7B8388}
\definecolor{DiagramEncoderFill}{HTML}{F1F3F4}
\definecolor{DiagramPredictor}{HTML}{4A7595}
\definecolor{DiagramPredictorFill}{HTML}{E8F0F5}
\colorlet{DiagramState}{DiagramPredictor}
\definecolor{DiagramAction}{HTML}{B67850}
\definecolor{DiagramActionFill}{HTML}{FBF1E9}
\definecolor{DiagramActionInk}{HTML}{8F5B3B}
\definecolor{DiagramLoss}{HTML}{5D7D6C}
\colorlet{DiagramLossFill}{FigureWhite}
\colorlet{ConceptAccent}{DiagramLoss}
\colorlet{DiagramContextFill}{FigureVisible}
\definecolor{DiagramStateCell}{HTML}{DCE8EF}
\definecolor{DiagramActionCell}{HTML}{F3E3D8}
\colorlet{LeWAMNavy}{LeWAMState}
\colorlet{LeWAMOrange}{LeWAMAction}
\colorlet{LeWAMGreen}{LeWAMObjective}
\colorlet{LeWAMGray}{FigureMuted}
\colorlet{LeWAMTeal}{PaperGray}
\colorlet{MethodLeWAM}{PaperBlue}
\colorlet{TableRule}{MethodLeWAM}
\colorlet{MethodLeWM}{PaperGray}
\colorlet{MethodDreamZero}{PaperGray}
\colorlet{MethodDINOWM}{PaperGray}
\colorlet{MethodPLDM}{PaperGray}
\colorlet{MethodGCBC}{PaperGray}
\colorlet{MethodGCIDM}{PaperGray}
\colorlet{MethodGCIDMScratch}{PaperGray}
\colorlet{MethodVJEPAAC}{PaperGray}
\colorlet{MethodSMWM}{PaperGray}
\colorlet{MethodRandom}{PaperGray}
\colorlet{MethodDiffusionPolicy}{PaperGray}
\colorlet{MethodACT}{PaperGray}

\arrayrulecolor{TableRule}
\usetikzlibrary{arrows.meta}
\tikzset{
  paper arrow/.style={-{Straight Barb[length=5pt,width=6pt]},
                      line width=0.9pt,line cap=round,line join=round},
  paper shaft/.style={-,line width=0.9pt,line cap=round,line join=round}
}

\DeclareRobustCommand{\endtoendstar}{%
  \tikz[baseline=-0.65ex,x=1.2em,y=1.2em]{%
    \filldraw[fill=PaperGrayDark,draw=FigureWhite,
              line width=0.04em,line join=bevel]
      (0,0.5) -- (-0.112256991,0.154508493)
      -- (-0.475528258,0.154508497) -- (-0.181635627,-0.059016993)
      -- (-0.293892626,-0.404508497) -- (0,-0.190983)
      -- (0.293892626,-0.404508497) -- (0.181635627,-0.059016993)
      -- (0.475528258,0.154508497) -- (0.112256991,0.154508493)
      -- cycle;
  }%
}

\newtcolorbox{reviewquestion}{
  enhanced,
  colback=gray!4,
  colframe=gray!35,
  boxrule=0.35pt,
  arc=1.5pt,
  left=6pt,
  right=6pt,
  top=5pt,
  bottom=5pt,
  before skip=7pt,
  after skip=5pt
}
\hypersetup{
  colorlinks=true,
  citecolor=LeWAMGray,
  linkcolor=black,
  urlcolor=black,
  pdfauthor={Minghao Fu, Tavis Siebert, Eryk Halicki, Randall Balestriero},
  pdftitle={LeWAM: End-to-End World and Action Modeling with JEPAs}
}

\newcommand{\methodname}[1]{\textup{#1}}
\newcommand{\dataset}[1]{\texttt{#1}}
\newcommand{\method}{\methodname{LeWAM}}
\newcommand{\methodintro}{\textbf{\textup{LeWAM}}}

\newcommand{\lead}[1]{\textbf{#1}}
\newcommand{\eg}{\textit{e.g.}\xspace}

\newcommand{\methoddot}[1]{%
  \tikz[baseline=-0.55ex]
    \filldraw[fill=#1, draw=black!75, line width=0.25pt]
      (0,0) circle[radius=0.38ex];%
}
\newcommand{\baselineitem}[2]{%
  \par\noindent\hangindent=0.62em\hangafter=1
  \makebox[0.62em][l]{\methoddot{#1}}#2\par
}
\newenvironment{compactbaselines}{%
  \par\begingroup\setlength{\parskip}{0pt}%
}{%
  \par\endgroup
}
\newcommand{\appendixbadge}[1]{%
  \tikz[baseline=(badge.base)]
    \node[circle, fill=gray!12, draw=gray!55, line width=0.35pt,
          text=gray!40!black, minimum size=1.45em,
          font=\bfseries\sffamily\footnotesize, inner sep=0pt] (badge) {#1};%
}
\newcommand{\projectbutton}[3]{%
  \href{#1}{%
    \tcbox[on line, colback=LeWAMStateWash, colframe=LeWAMStateWash, coltext=MethodLeWAM,
           boxrule=0.4pt, arc=4pt, boxsep=2pt,
           left=4pt, right=4pt, top=2pt, bottom=2pt]{%
      \sffamily\mdseries\upshape\small #2\enspace #3}}%
}

\title{LeWAM: End-to-End World and Action \\Modeling with JEPAs}

\author{%
Minghao Fu\thanks{Equal contribution. Correspondence to \href{mailto:m9fu@ucsd.edu}{\texttt{m9fu@ucsd.edu}}.}\kern0.42em$^{1}$\quad
Tavis Siebert$^{*2}$\quad
Eryk Halicki$^{3}$\quad
Randall Balestriero$^{4}$\\[-0.1ex]
{\normalfont\small
$^{1}$UC San Diego\quad
$^{2}$ETH Zurich\quad
$^{3}$University of British Columbia\quad
$^{4}$Brown University
}
}

\iclrfinalcopy
\begin{document}

\maketitle
\lhead{Preprint}
\vspace{-0.30in}
{\centering
\projectbutton{https://le-wam.github.io/}{\faGlobe}{Website}\hspace{0.5em}%
\projectbutton{https://github.com/MinghaoFu/lewam}{\faGithub}{Code}\par}
\vspace{3pt}

\begin{abstract}
Joint-embedding predictive architectures (JEPAs) learn compact latent world models without pixel reconstruction. Existing JEPA pipelines typically train visual representations and action-conditioned dynamics before adding a control policy or solver. We introduce \methodintro{}, an end-to-end JEPA for control with: (1) a joint training objective over next latent state prediction and action flow matching, and (2) an isotropic Gaussian constraint on the latent state, enabling a single model to support direct policy execution and efficient planning by evaluating sampled actions with its learned latent dynamics. Compared with prior JEPAs in visual planning, LeWAM improves success rate from 28.6\% to 89.7\% on contact-rich tasks and from 10.9\% to 46.2\% on long-horizon tasks, and plans $32.3\times$ faster. On behavior cloning tasks in simulation and physical robots, LeWAM is also competitive with established methods. Interpretability analyses show that LeWAM learns control-relevant representations and action distributions effectively guiding test-time planning and control.
\end{abstract}\vspace{-6.5pt}

\FloatBarrier
\begin{figure}[H]
\centering
\includegraphics[width=\linewidth,trim=0 31.68 0 83.52,clip]{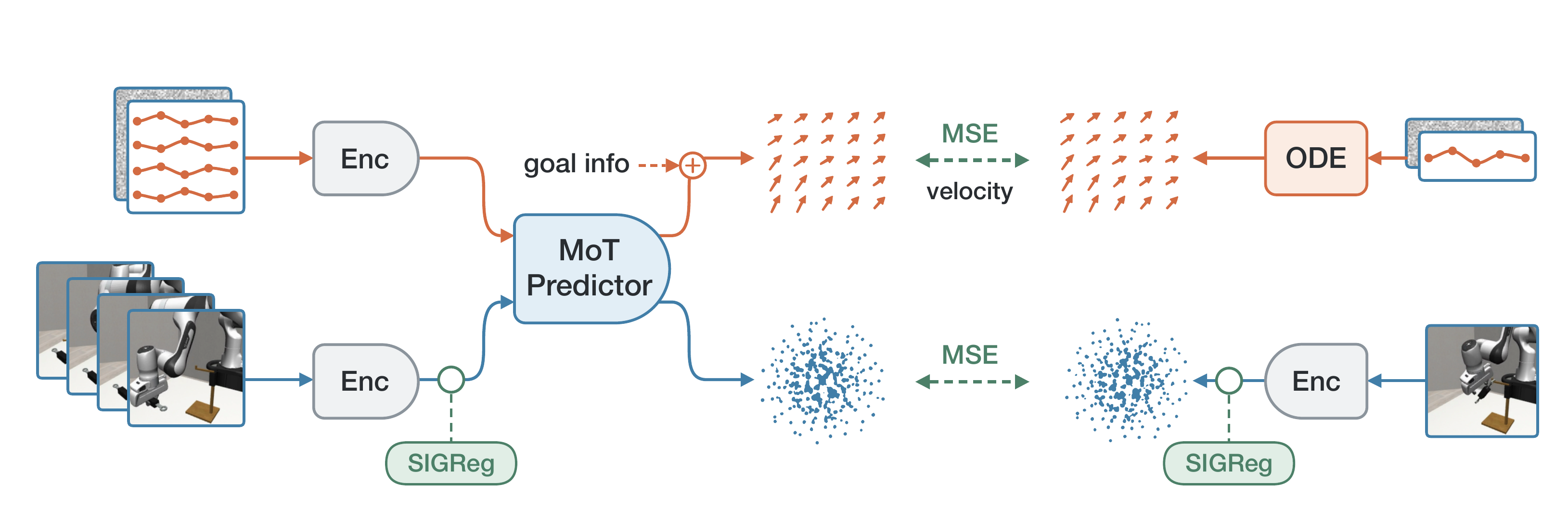}
\par\vspace{0pt}
{\color{LeWAMGray!40}\rule{\linewidth}{0.3pt}}
\par\vspace{0pt}
\includegraphics[width=\linewidth]{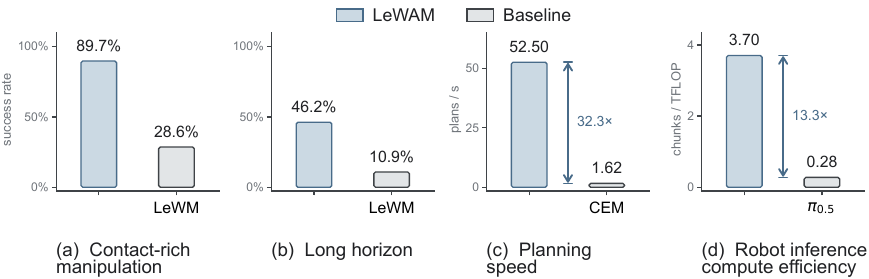}
\setlength{\abovecaptionskip}{1pt}
\caption{\lead{LeWAM Training.} A shared encoder encodes observed frames with 
\methodname{SIGReg}~\citep{balestriero2025lejepa} regularization.
The Mixture-of-Transformers~\citep{liang2025MoT} learns latent-state prediction
from clean actions and action flow matching from noisy actions with goal information.
\lead{Results Overview.} The mean of (a) planning success on contact-rich tasks; (b) planning success on a 100-step goal offset; (c) plans/s versus CEM~\citep{rubinstein1999cem}; (d) action chunks
per TFLOP versus $\pi_{0.5}$~\citep{physicalintelligence2025pi05}. Full results are in Section~\ref{sec:experiments}.}
\label{fig:intro}
\label{fig:headline-results}
\end{figure}

\clearpage
\begin{figure}[!t]
\centering
\includegraphics[width=\linewidth]{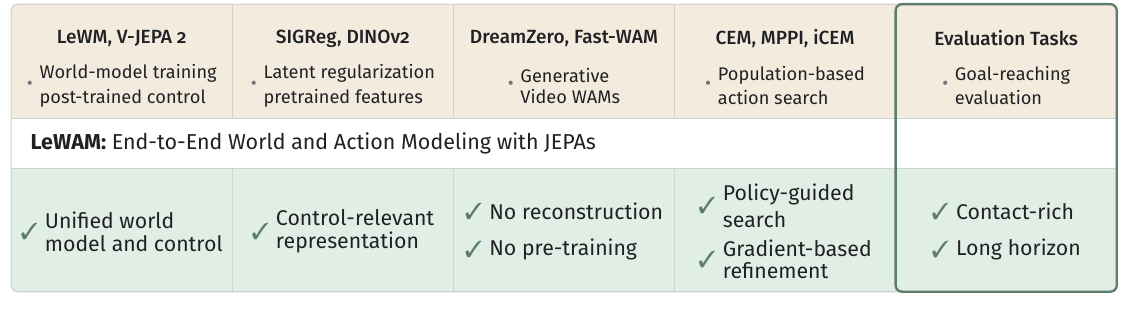}
\vspace{-2em}
\caption{\lead{Positioning of LeWAM.}
The columns place \method{} alongside related approaches to world-model training, visual representation learning, world action modeling, and action search. \method{} jointly trains latent dynamics and a policy through a shared encoder, predicts in latent space without reconstruction, and refines learned action proposals through its dynamics. The final column summarizes its evaluation on contact-rich tasks and longer goal horizons.}
\label{fig:three-consequences}
\vspace{-1em}
\end{figure}

\section{Introduction}
\label{sec:introduction}

Visual control requires autonomous agents to complete tasks by selecting actions based on visual observations of the world. World models address this challenge by predicting the consequences of candidate actions, allowing agents to anticipate how the world will evolve before executing them. Joint-embedding predictive architectures (JEPAs)~\citep{lecun2022apath} offer a compact world modeling paradigm where future world states are predicted in latent space rather than by reconstructing pixels. 
Recent works such as \methodname{DINO-WM}~\citep{zhou2024dinowm} and \methodname{LeWorldModel (LeWM)}~\citep{maes2026lewm} demonstrate the promise of applying action-conditioned latent world models to downstream control. These pipelines obtain visual features from pretrained encoders~\citep{zhou2024dinowm,assran2025vjepa2}, or train representations from scratch with distributional regularizers such as \methodname{SIGReg}~\citep{balestriero2025lejepa,maes2026lewm} to prevent collapse. However, these methods alone do not reliably retain control-relevant information; JEPAs can favor temporally persistent distractors~\citep{sobal2022slowfeatures}, while action-conditioned dynamics may retain extraneous observation information~\citep{brandfonbrener2023inverse}.

Control with these models typically requires an online solver such as the cross-entropy method (CEM)~\citep{rubinstein1999cem} or a separately trained action predictor. When initialized with a broad Gaussian prior over action sequences, CEM demands extensive rollouts to identify viable candidates, particularly as action dimensionality or planning horizon grows~\citep{huang2021cemgd}. Contact transitions also complicate local action optimization because the dynamics can switch between modes~\citep{pang2023contactplanning}. A post-trained action predictor can provide direct control, as in \methodname{GC-IDM}~\citep{nguyen2026gcidm}, or guide action search, as in \methodname{PRISM}~\citep{wang2026prism}. Training these predictors on frozen JEPA embeddings leaves action supervision disconnected from visual representation learning.

We introduce \methodintro{}, an end-to-end JEPA that jointly learns
representation, action-conditioned dynamics, and a stochastic policy
in a single training stage (Figure~\ref{fig:intro}). Training combines two ingredients: a joint training objective over next-latent-state prediction and action flow matching, and \methodname{SIGReg}~\citep{balestriero2025lejepa}
to regularize the latent states. A Mixture-of-Transformers (MoT)~\citep{liang2025MoT} predictor uses clean actions for latent dynamics prediction and noisy actions for policy learning through flow matching, so both objectives train the shared visual representation. Consequently, LeWAM supports both direct policy execution and visual planning by refining action sequences sampled from its learned policy. Our contributions are threefold:
\begin{itemize}
    \item To our knowledge, the first JEPA that jointly learns representation, action-conditioned latent dynamics, and stochastic policy, using a joint prediction objective, one weight for latent-space regularization, and approximately 17M parameters.
    
    \item A planning procedure that jointly imagines the consequences of actions and how to act in the resulting states, which samples candidate action sequences from the learned action distribution and refines them with gradient descent through learned dynamics.

    \item Across ten heterogeneous simulated environments, \method{} improves goal-reaching success over \methodname{LeWM}~\citep{maes2026lewm}, particularly on contact-rich tasks. Four-task comparisons extend these gains to longer goal horizons and show faster planning. On goal-agnostic task completion in simulation and physical robot execution, \method{} performs competitively with established baselines. Linear probes reveal decodable physical-state and action information, recorded-action rollouts visualize its learned dynamics, and planner ablations measure the gains from its stochastic action proposals.
\end{itemize}

\section{Related Work}
\label{sec:related}
\lead{World Models.} World models~\citep{ha2018worldmodels} learn environment dynamics to enable agents to simulate candidate action outcomes for planning and control. Early approaches like \methodname{PlaNet}~\citep{hafner2019planet} perform planning via latent rollouts, whereas \methodname{TD-MPC}~\citep{hansen2022tdmpc} and \methodname{TD-MPC2}~\citep{hansen2024tdmpc2} ground latent dynamics in reward and value predictions. \methodname{Dreamer}~\citep{hafner2020dreamer,hafner2023dreamerv3} learns generative latent dynamics with pixel reconstruction, while \methodname{VideoGPT}~\citep{yan2021videogpt} and \methodname{IRIS}~\citep{micheli2023iris} model discrete visual tokens. \method{} learns latent dynamics and a policy directly from offline trajectories, without reward supervision or pixel reconstruction.

\lead{Joint-Embedding Predictive Architectures.} JEPAs predict target states in representation space without pixel
reconstruction, using variance and covariance constraints~\citep{bardes2022vicreg}, momentum
targets~\citep{assran2023ijepa,bardes2024vjepa}, or distributional
regularization~\citep{balestriero2025lejepa,wu2026visreg} to prevent collapse.
For action-conditioned world modeling,
\methodname{DINO-WM}~\citep{zhou2024dinowm} and \methodname{V-JEPA 2-AC}~\citep{assran2025vjepa2}
build on pretrained representations, whereas \methodname{LeWM}~\citep{maes2026lewm} leverages
SIGReg~\citep{balestriero2025lejepa} to learn the encoder and latent dynamics from scratch.
Additional action prediction objectives attempt to make representations action-aware in \methodname{PLDM}~\citep{sobal2025pldm} or as an anti-collapse signal in
\methodname{SMWM}~\citep{ivashkov2026smwm}. \method{} jointly trains action-conditioned latent
dynamics and policy in one predictor, so both consequence prediction and action supervision
shape the shared representation.

\lead{Control with Latent Dynamics.} Conventional planning in latent world models uses population-based solvers, such as CEM~\citep{rubinstein1999cem}, iCEM~\citep{pinneri2021icem}, MPPI~\citep{williams2017mppi}, or gradient-based methods~\citep{henaff2017discrete,huang2021cemgd}. Population-based search can require many model rollouts as the action dimension or horizon grows~\citep{pinneri2021icem,huang2021cemgd}, and contact mode changes in manipulations also complicate local optimization~\citep{pang2023contactplanning} (more discussion in Appendix~\ref{app:joint-prediction}). \methodname{GC-IDM}~\citep{nguyen2026gcidm} post-trains a goal-conditioned action predictor for direct control, whereas \methodname{PRISM}~\citep{wang2026prism} trains a Gaussian prior head on frozen world-model features; its predicted mean and standard deviation guide MPPI sampling. \method{} jointly trains the latent world model and a flow-based action predictor, using its sampled actions for direct control or gradient-based planning.

\lead{World Action Models.} These models couple world prediction with action prediction
(Table~\ref{tab:world-action-factorizations}, Appendix~\ref{app:system-simplicity}).
Built on pretrained video generative models, \methodname{DreamZero}~\citep{ye2026dreamzero} jointly models future video and actions, while \methodname{Fast-WAM}~\citep{yuan2026fastwam} retains video co-training but skips future video generation at inference. \methodname{UVA}~\citep{li2025uva}, \methodname{UWM}~\citep{zhu2025uwm}, and \methodname{Cosmos 3}~\citep{nvidia2026cosmos3} specifically support both forward dynamics modeling (FDM) and policy learning, but optimize only one objective per sample, rather than both jointly on the same pass. \method{} jointly trains its encoder and predictor end-to-end using latent-state prediction and action flow matching, without pixel reconstruction. Each training sample supplies
clean actions for FDM and noisy actions for stochastic policy modeling. \method{} learns action-conditioned dynamics from expert and non-expert interactions and uses them to evaluate and refine actions at test time.
\vspace{-1em}

\section{Methodology}
\label{sec:method}

Given reward-free offline trajectories of pixel observations $\vo_{1:T}$ and actions $\va_{1:T}$, we jointly train a visual encoder and a predictor including action-conditioned dynamics and a flow matching policy, supporting direct control and planning with action-conditioned latent dynamics.

\subsection{End-to-End Training with LeWAM}
\label{sec:joint-predictor}

\Needspace{20\baselineskip}
\begin{wrapfigure}{r}{0.40\textwidth}
\centering
\includegraphics[width=\linewidth]{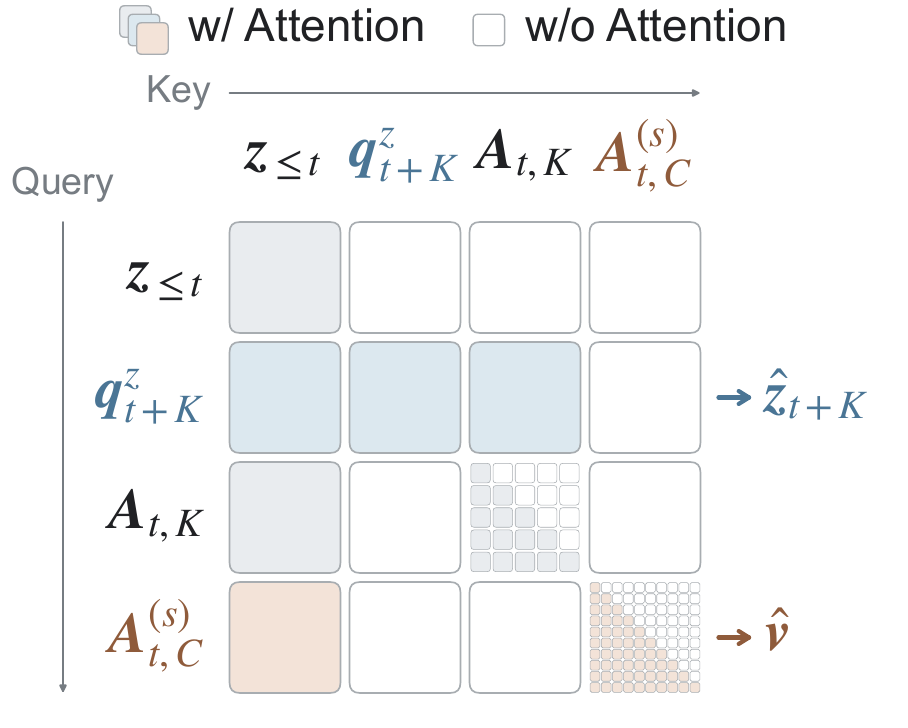}
\vspace{-1em}
\setlength{\abovecaptionskip}{4pt}
\small
\caption{\lead{MoT Attention.} Filled cells permit
attention, while white cells are masked. The state
query reads history and clean actions, and noisy action tokens read state history and their
own causal prefix. Small grids show causal attention. Arrows mark the
state branch and action velocity branch.}
\label{fig:mot-architecture}
\vspace{-2em}
\end{wrapfigure}

\lead{Architecture.} Figure~\ref{fig:intro} summarizes the model at a high level. Let $\vz_t=\enc(\vo_t)$ denote the latent representation of observations $\vo_t$. All time indices refer to raw environment steps, such that action $\va_t$ transitions $\vo_t$ to $\vo_{t+1}$. We use a frameskip of $K$ environment steps, so future observations used as dynamics targets are spaced at $K$-step intervals. In contrast, the policy models a contiguous action chunk $\mA_{t,C} = (\va_t,\dots,\va_{t+C-1})$ with $C\geq K$. For visual planning, we assume access to goal information $\vc_t$ that contains the goal image embedding $\vz_g$ and its normalized remaining horizon from $\vz_t$.

The encoder consists of a ResNet-18 backbone~\citep{he2016resnet}, spatial softmax pooling, and a multilayer perceptron (MLP) with two projection layers. The same encoder processes all the pixel observations: history, future target, and goal image. The predictor is comprised of a state branch $\statebranch$ and an action branch $\actionbranch$. Specifically,
\begin{equation}
    \begin{aligned}
        \hat{\vz}_{t+K}
        &=\statebranch(\vq^z_{t+K};\vz_{\leq t},\mA_{t,K}),\\
        \hat{\vv}_{t,C} &=\actionbranch(\mA_{t,C}^{(s)};s,\vz_{\leq t},\vc_t)
    \end{aligned}
    \label{eq:lewam-interface}
\end{equation}
where $\mA_{t,C}^{(s)}$ is a noisy $C$-action chunk at flow timestep $s\in[0,1]$ (see Equation~\ref{eq:action-flow-path}) and $\vq^z_{t+K}$ is a learned query token specifying the future state to predict. Integrating the predicted velocity field $\hat{\vv}_{t,C}$ yields the predicted action chunk $\hat{\mA}_{t,C}$.

To simultaneously balance latent-prediction and flow-matching objectives while allowing $\statebranch$ and $\actionbranch$ to share information about states and actions, we use the Mixture-of-Transformers (MoT)~\citep{liang2025MoT} architecture. That is, each branch applies independent query, key, value, and feedforward projections for its respective modality (latent states or actions) while sharing one masked self-attention. Figure~\ref{fig:mot-architecture} specifies the information available to each branch during attention. The state query $\vq^z_{t+K}$ reads history latents and clean actions, and noisy action tokens $\mA_{t,C}^{(s)}$ read history latents and their own causal prefix. History latents attend only to themselves, blocking indirect clean-action leakage to $\actionbranch$.
Goal information conditions the noisy action tokens; the dynamics branch reads history latents and clean actions.
Appendix~\ref{app:implementation} gives the full token-level mask, layer sizes, and training settings.
\finishwrapfigure 

\lead{Training Objective.} For flow matching~\citep{lipman2023flowmatching}, given a recorded action chunk $\mA_{t,C}$,
sample $s$ uniformly from $[0,1]$ and
$\vepsilon\sim\mathcal{N}(\vzero,\mI)$, then construct
\begin{equation}
    \mA^{(s)}_{t,C}=(1-s)\vepsilon+s\mA_{t,C},
    \qquad \vv_{t,C}=\frac{\mathrm d\mA^{(s)}_{t,C}}{\mathrm ds}=\mA_{t,C}-\vepsilon
    \label{eq:action-flow-path}
\end{equation}
The action branch regresses the path velocity $\mA_{t,C}-\vepsilon$ from $\mA_{t,C}^{(s)}$.
The latent dynamics and action flow matching losses use mean squared error (MSE), averaged over the scalar coordinates of each prediction:
\begin{equation}
    \begin{aligned}
    \gL_{\mathrm{dyn}}
    &=\mathbb{E}_{\mathcal D}\!\left[
        \operatorname{MSE}\!\left(
        \hat{\vz}_{t+K},\vz_{t+K}\right)\right], \\
    \gL_{\mathrm{act}}
    &=\mathbb{E}_{\mathcal D,s,\vepsilon}\!\left[
        \operatorname{MSE}\!\left(
        \hat{\vv}_{t,C},\mA_{t,C}-\vepsilon
        \right)\right].
    \end{aligned}
    \label{eq:prediction-losses}
\end{equation}
We use
\methodname{SIGReg}~\citep{balestriero2025lejepa} to preserve latent diversity
through an isotropic Gaussian distributional constraint. Let $\mZ$ collect
the encoded history and target observations in a minibatch. With $M$ random unit
directions $\vu^{(m)}$ and the univariate Epps--Pulley~\citep{epps1983test}
test statistic $T$ against $\mathcal N(0,1)$, the regularizer is
\begin{equation}
    \gL_{\mathrm{sig}}
    =\frac{1}{M}\sum_{m=1}^{M}
    T\!\left(\mZ\vu^{(m)}\right).
    \label{eq:sigreg-objective}
\end{equation}
With regularization weight $\lambda_{\mathrm{sig}}$, the full objective is
\begin{equation}
    \gL_{\rm LeWAM}
    =\gL_{\mathrm{dyn}}+\gL_{\mathrm{act}}
    +\lambda_{\mathrm{sig}}\gL_{\mathrm{sig}}.
    \label{eq:joint-objective}
\end{equation}

\begin{figure}[!t]
\centering
\includegraphics[width=\linewidth,trim=0 94 0 20,clip]{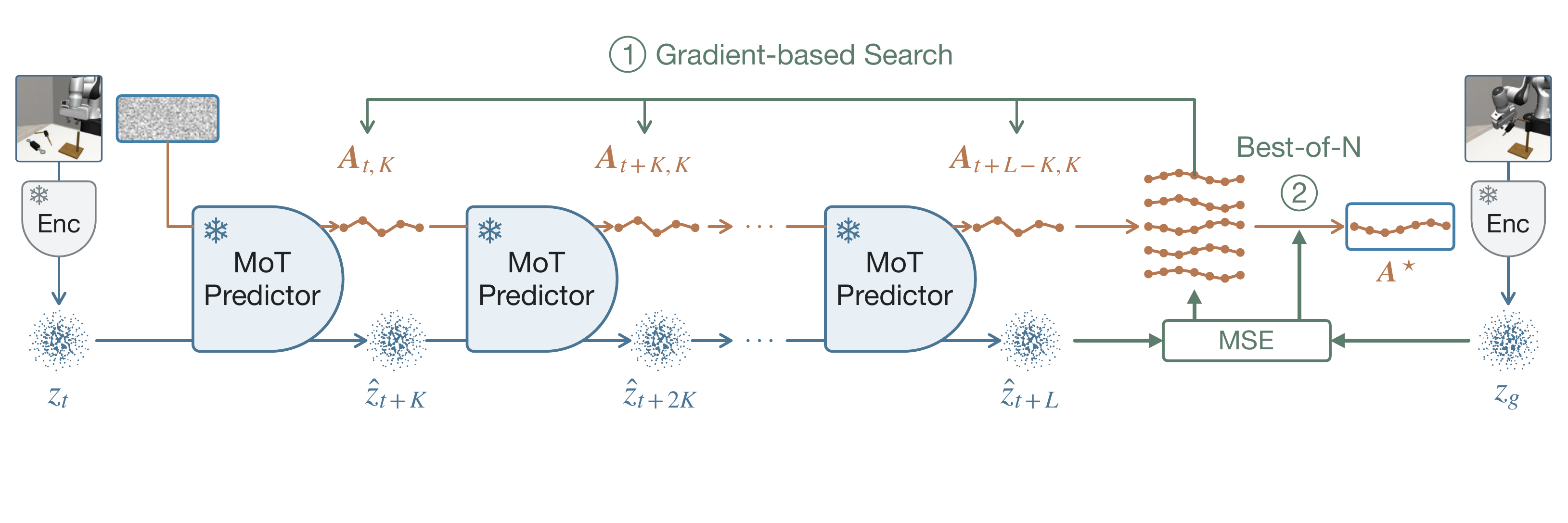}
\vspace{-2em}
\caption{\lead{Planning with LeWAM.}
Each MoT predictor stage samples $N$ action chunks from Gaussian noise and uses the first $K$ actions in each chunk, denoted $\mA_{t,K}$,
to advance the latent dynamics from step $t$ to $t+K$.
Five-point curves depict action chunks (we use $K=5$), and stacked curves denote $N$ sampled plans of length $L$. Step 1 refines all plans by gradient descent through the frozen LeWAM dynamics using the goal-state cost. Step 2 selects the lowest-cost refined iterate as the final plan, shown as $\mA^\star$. Appendix~\ref{app:planners} gives the full procedure in a pseudo-algorithm.}
\label{fig:lewam-planning}
\end{figure}

\Needspace{12\baselineskip}
\subsection{Efficient Control with LeWAM}
\label{sec:control-planning}

\method{} supports both direct control through its learned policy and planning through rollouts that alternate action sampling and latent-state prediction.

\lead{Direct Control.} With history and optional goal context fixed, one $C$-action chunk is sampled by integrating along the learned velocity field over the interval $[0,1]$ from Gaussian noise:
\begin{equation}
    \frac{\mathrm d\mA_{t,C}^{(s)}}{\mathrm ds}
    =\actionbranch(\mA_{t,C}^{(s)};s,\vz_{\leq t},\vc_t),
    \qquad \mA_{t,C}^{(0)}\sim\mathcal{N}(\vzero,\mI),
    \qquad \hat{\mA}_{t,C}=\mA_{t,C}^{(1)}
    \label{eq:action-initialization}
\end{equation}
The first $K$ actions are executed and resampling begins from the resulting observation.

\lead{Planning.} \method{} uses a gradient-based refinement strategy over a population of action candidates sampled from the learned action distribution (Figure~\ref{fig:lewam-planning}).
To simplify notation, let $\mA=(\va_t,\dots,\va_{t+L-1})$ denote a plan to reach the goal $\vz_g$ after $L$ raw environment steps from time $t$, where $L$ is a multiple of frameskip $K$.

Starting from the observed history $\vz_{\leq t}$, one population of $N$ candidate plans is generated from the learned action distribution by alternating action sampling and latent dynamics. Specifically, at each step, Eq.~(\ref{eq:action-initialization}) generates $N$ action chunks from the observation history and goal. The forward dynamics branch uses
the first $K$ actions of each chunk to predict the $N$ corresponding next states.
Each predicted state is appended to the fixed-length history of its candidate, the planning horizon is updated, and the process continues for $L$ total steps. This leaves $N$ action sequence candidates $\{\mA^{(n)}\}_{n=1}^N$. 
Starting from $\{\mA^{(n)}\}_{n=1}^N$, we refine each candidate by gradient descent through latent dynamics~\citep{henaff2017discrete,jyothir2023gradient} with respect to the latent cost-to-goal
\begin{equation}
     \gC(\mA)
     =\operatorname{MSE}\!\left(\hat{\vz}_{t+L}(\mA),\vz_g\right)
     \label{eq:lewam-gradient-planning}
\end{equation}
With learning rate $\eta$, the goal-cost update takes the gradient-descent form
\begin{equation}
    \va_{t+k}\leftarrow\va_{t+k}-\eta\frac{\partial\gC}{\partial\va_{t+k}},
    \qquad k=0,\ldots,L-1,
    \label{eq:action-gradient-refinement}
\end{equation}
The descent iterate with the lowest $\gC$ across all candidates is chosen. In practice, we add a small proximity penalty to discourage large changes from the sampled plans (Appendix~\ref{app:planners}). We execute the first $K$ actions, observe the resulting state, and repeat sampling and refinement until the goal is reached, as in receding-horizon model predictive control.

\section{Experiments}
\label{sec:what-joint-buys}
\label{sec:experiments}

Our experiments aim to address the following questions:
(1)~Can a single LeWAM architecture learn stably from offline data and support visual planning across heterogeneous tasks and settings?
(2)~Does LeWAM jointly learn informative action-conditioned latent dynamics, control-relevant representations, and multi-modal distributions over actions?
(3)~Can LeWAM further support behavior cloning in simulated environments and physical-robot control?




\subsection{Evaluation Protocol}
\label{sec:worldforge8}
\begin{figure}[!t]
\centering
\includegraphics[width=0.9\linewidth]{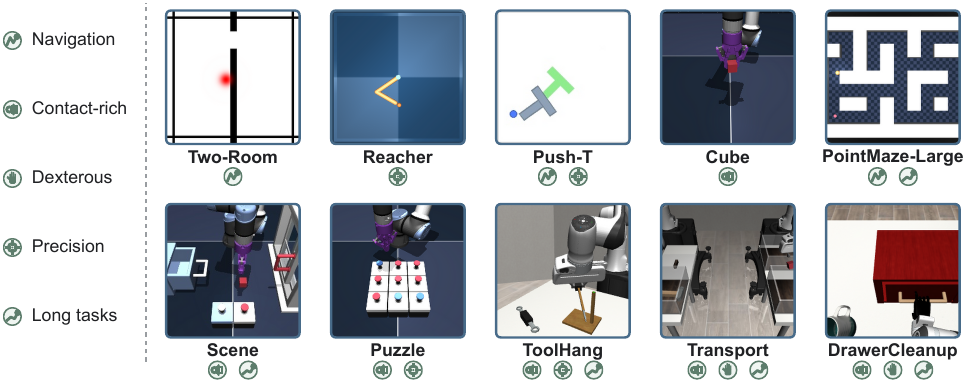}
\vspace{-1em}
\caption{\lead{Simulation Environments.} Symbols beneath each frame denote overlapping task attributes,
with their meanings shown in the legend to the left.}
\label{fig:worldforge8}
\vspace{-1em}
\end{figure}

\lead{Heterogeneous Simulated Environments} We evaluate control across ten simulated navigation and manipulation
environments (Figure~\ref{fig:worldforge8}). The environments vary in task functionality and temporal horizon, and their
datasets include expert demonstrations, random-policy interactions, and mixed-quality data. Four environments adapt the benchmarks used in
\methodname{LeWM}~\citep{maes2026lewm}. The other six add challenging contact-rich manipulation and navigation. Appendix~\ref{app:benchmark-evaluation} details task construction, data, and success criteria. For \dataset{Cube}, success checks both cube and end-effector positions.

\lead{Evaluation Protocols.} We distinguish two separate evaluation settings. Visual planning assumes access to a goal image, and a goal-conditioned policy or planner must reach the depicted state from its current (random) starting position within a fixed step budget. Task completion provides no goal image and evaluates behavior cloning from demonstrations: policies execute actions from start to finish to complete one specific task. Within each setting,
methods share training and evaluation data, observations, action limits, goal offsets, execution budgets, and success predicates.
Appendix~\ref{app:splits} details goal construction, evaluation conditions, and success criteria.

\subsection{Visual Planning with LeWAM}
\label{sec:goal-conditioned-results}
We compare with four goal-conditioned baselines spanning latent-dynamics
planning, an inverse-dynamics policy on a frozen world model, and direct policy execution:
\begin{compactbaselines}
\baselineitem{MethodLeWM}{\textbf{\methodname{LeWM (CEM)}}~\citep{maes2026lewm}. Learning compact action-conditioned latent dynamics end-to-end with SIGReg, planning with CEM over learned latent dynamics.}
\baselineitem{MethodGCIDM}{\textbf{\methodname{LeWM (IDM)}}~\citep{nguyen2026gcidm}. Post-trains a goal-conditioned IDM on frozen LeWM.}
\baselineitem{MethodDINOWM}{\textbf{\methodname{DINO-WM}}~\citep{zhou2024dinowm}. Learns dynamics on a frozen DINOv2 encoder with CEM planner.}
\baselineitem{MethodGCBC}{\textbf{\methodname{GCBC}}~\citep{ghosh2021gcsl}. Trains a goal-conditioned policy on frozen DINOv2 features.}

\end{compactbaselines}

\begin{figure}[t]
\centering
\includegraphics[width=\linewidth]{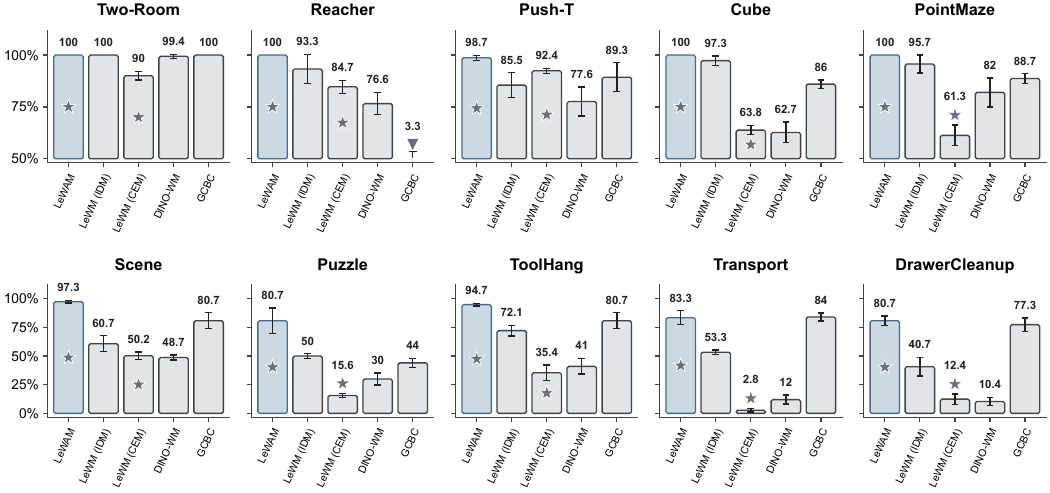}
\vspace{-2em}
\caption{\lead{Visual Planning across Ten Simulated Environments.} Most panels report the success rate under a goal offset of $L=25$ environment steps and permit $2L=50$ actions across 3 seeds. For \dataset{Two-Room}, we use 100/150 action steps following LeWM setting; For LeWM (CEM), we report 10 seeds for stability. Details are present in Appendix~\ref{app:splits}.
Error bars denote one standard deviation; \endtoendstar{} indicates models with end-to-end training. Linear axes start at 50\% for the first five panels.}
\label{fig:short-horizon-control}
\end{figure}

Figure~\ref{fig:short-horizon-control} reports goal-reaching success across ten environments. Across the ten tasks, \method{} averages $93.5\%$ success versus $50.9\%$ for
\methodname{LeWM (CEM)}~\citep{maes2026lewm}, a $42.7$ percentage-point gap.
\method{} exceeds that baseline on every task, with particularly large gaps on
contact-rich \dataset{ToolHang} ($94.7$ versus $35.4$) and
\dataset{Transport} ($83.3$ versus $2.8$).
On the logically structured \dataset{Puzzle} task,
\method{} reaches $80.7\%$ success versus $44.0\%$ for goal-conditioned \methodname{GCBC}~\citep{ghosh2021gcsl}.

\begin{figure}[!htbp]
\centering
\vspace{-1em}
\includegraphics[width=\linewidth]{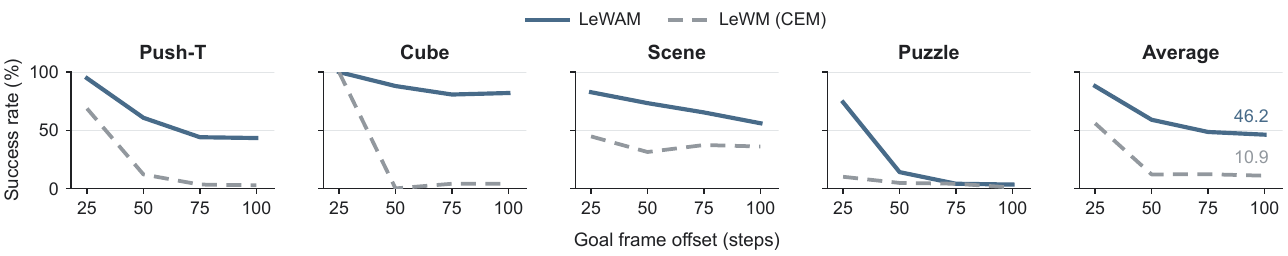}
\vspace{-1em}
\caption{\lead{Visual Planning Success across Different Horizons.}
The task panels show mean success rates across 3 seeds. Each sweep starts at the first trajectory frame but with different goal offsets.}
\label{fig:planning-horizon}
\vspace{-1em}
\end{figure}

\label{sec:long-horizon-planning}
\lead{Long-Horizon Planning.} Using the same checkpoints, we move the goal farther along each recorded trajectory while keeping its starting frame fixed on \dataset{Push-T}, \dataset{Cube}, \dataset{Scene}, and \dataset{Puzzle}. Unlike navigation, these tasks introduce significantly higher behavioral diversity as the planning horizon grows. At a 100-step goal offset, \method{} achieves $46.2\%$ mean success across these four tasks versus $10.9\%$ for LeWM (Figure~\ref{fig:planning-horizon}).

\label{sec:planning-efficiency}
\lead{Towards Scalable Planning.} Beyond accuracy gains, \method{} delivers planning acceleration through two core mechanisms: (1) an expressive distribution prior that narrows the search space, and (2) gradient-based optimization that leverages GPU's parallel architectures to unlock scalable test-time search. Across four visual-goal tasks, it averages $52.50\text{ plans/s}$ versus $1.62\text{ plans/s}$ for \methodname{LeWM (CEM)}, a $32.3\times$ speedup (Table~\ref{tab:planning-speed}; Figure~\ref{fig:headline-results}(c)).

\subsection{Understanding LeWAM's Dynamics, Representations, and Actions}
\label{sec:world-dynamics}
\label{sec:physical-understanding}




\lead{Latent Representation Probing.} To better measure what information is learned by our joint-training objective, we evaluate how well linear probes recover physical states, proprioceptive information, and actions from \method{}'s representations. As highlighted in Table~\ref{tab:physical-probing-main} and Appendix~\ref{app:latent-probing}, training with an action modeling objective noticeably improves LeWAM's linear action and proprioception decodability compared to LeWM, increasing average proprioception, action, and IDM $R^2$ by $+0.21, +0.20, +0.19$ respectively (see Table~\ref{tab:physical-probing-appendix}, Appendix~\ref{app:latent-probing}).

\lead{The Role of SIGReg.} SIGReg~\citep{balestriero2025lejepa,maes2026lewm} constrains the latent distribution to prevent representation collapse. Without it, latent-only LeWM yields $R^2=0.00$ on every reported \dataset{Push-T} probe, while LeWAM with either action head retains decodable state and action information (Table~\ref{tab:physical-probing-main}). This supports action supervision as an \emph{anti-collapse signal}: observed actions anchor the learned representation to targets independent of the encoder.
However, only constraints from action prediction is not enough, beyond anti-collapse, we find that SIGReg can largely improve planning: on \dataset{Push-T}, the planning gain is 14.7 percentage points with SIGReg, compared to 2.7 without it (Table~\ref{tab:planning-component-ablation}). The probes in Tables~\ref{tab:physical-probing-main} and~\ref{tab:physical-probing-appendix} show that SIGReg improves object-state decodability as well. This is because SIGReg can retain informative latent representation and dynamics which help downstream planning, and make latent distance-to-goal more sensitive to changes in object state.
\begin{table}[t]
\centering
\caption{Linear probing and the role of SIGReg. All linear probes report $R^2$ score using ridge regression. \emph{Object} denotes the average $R^2$ over all object ground truth information and \emph{Proprio} the ground truth proprioceptive state. \emph{Action} predicts the action chunk from latent $\vz_t$; parenthesized IDM scores use both $\vz_t,\vz_{t+K}$ where reported. \emph{Eff. rank} denotes effective rank~\citep{royvetter2007effrank}.}
\label{tab:physical-probing-main}
\small
\setlength{\tabcolsep}{6pt}
\renewcommand{\arraystretch}{1.10}
\begin{tabular}{@{}lllcccc@{}}
\toprule
Environment & Model & SIGReg & Eff. rank & Object $\uparrow$ & Proprio $\uparrow$ & Action (IDM) $\uparrow$ \\
\midrule
\multirow{5}{*}{\dataset{Push-T}} & \multirow{2}{*}{\method{}} & Yes & $42.90$ & $0.90$ & $\mathbf{0.86}$ & $\mathbf{0.40} (\mathbf{0.75})$ \\
& & No & $10.94$  & $0.88$ & $0.85$ & $0.38 (0.73)$ \\
\cmidrule(l){2-7}
& \method{} (MSE) & No & $6.96$ & $0.66$ & $\mathbf{0.86}$ & $0.36\,(0.74)$ \\
\cmidrule(l){2-7}
& \multirow{2}{*}{\methodname{LeWM}} & Yes & $94.20$ & $\mathbf{0.95}$ & $0.80$ & $0.29 (0.71)$ \\
& & No & $16.23$ & $0.00$ & $0.00$ & $0.00\,(0.00)$ \\
\bottomrule
\end{tabular}
\vspace{-1em}
\end{table}

\begin{figure}[htbp!]
\centering
\includegraphics[width=\linewidth]{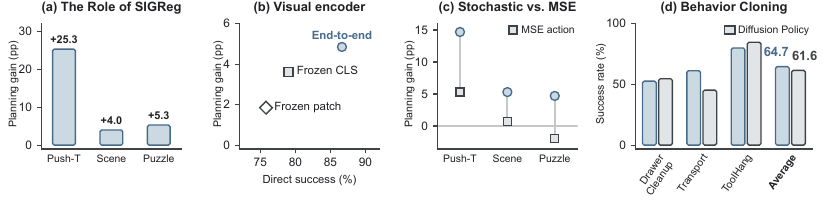}
\vspace{-2em}
\caption{
(a) Planning gain (percentage points) from SIGReg. (b) Direct success and planning gain for visual
encoders, averaged over four tasks. (c) Planning gain for flow (circles) and
MSE (squares) action heads.
(d) Task-completion success for LeWAM and
\methodname{Diffusion Policy}.}
\vspace{-1em}
\label{fig:model-ablations}
\end{figure}

Figure~\ref{fig:model-ablations}(a) compares
full \method{} with its no-SIGReg ablation. Removing SIGReg lowers planning
success in the three tasks shown, most strongly on \dataset{Push-T} (25.3 percentage
points), followed by \dataset{Puzzle} (5.3) and \dataset{Scene} (4.0).

\phantomsection
\label{sec:action-refines-representation}
\lead{Why Train the Visual Encoder End-to-End?} Frozen DINOv2~\citep{oquab2023dinov2} features underperform end-to-end features in our setup (Figure~\ref{fig:model-ablations}(b)). Across the four tasks, direct success averages $86.7\%$ with the jointly trained encoder, versus $79.0\%$ with frozen CLS features and $75.8\%$ with frozen patch features. Joint training thus improves the representation used for action selection; the additional gain from planning shows that its learned dynamics also support action refinement.
Appendix~\ref{app:simulation-rollouts} provides aligned visual reconstructions and
recorded-action open-loop rollouts for four simulated tasks.

\phantomsection
\label{sec:ablations}
\lead{Why Model Stochastic Action Distributions?} 
With the same latent-dynamics objective, flow matching yields larger planning gains than an MSE head whose point predictions are perturbed with Gaussian noise (Figure~\ref{fig:model-ablations}(c)). Without SIGReg, flow matching also retains more linearly decodable object information ($R^2=0.88$ versus $0.66$), with comparable proprioception scores ($0.85$ versus $0.86$). Deterministic MSE targets the conditional mean $\mathbb{E}[\mA_{t,C}\mid\vz_{\leq t},\vc_t]$, whereas flow matching models the conditional action distribution. It can therefore retain distinctions between states whose action distributions differ despite similar means. Appendix~\ref{app:action-distributions} illustrates this distinction in a two-door navigation experiment, where LeWAM presents a explicit multi-modal action distribution.

\paragraph{Isolating Action-Conditioned Dynamics.}
To examine the dynamics learned under \method{}'s joint objective, we remove its action prediction head at evaluation time and use CEM to plan with only its action-conditioned latent dynamics. Compared with \methodname{LeWM}, which trains action-conditioned dynamics without policy learning, \method{} produces comparable visual predictions and achieves competitive planning success on \dataset{Push-T}. On \dataset{Cube}, a 3D manipulation task, LeWAM has clearer visual predictions, especially in control-relevant feature, \eg, gripper and object, and higher planning success  (Appendix~\ref{app:simulation-rollouts}, Figure~\ref{fig:simulation-rollouts-pusht-cube}, and Table~\ref{tab:dynamics-only-cem}). These results show that joint action learning is compatible with world modeling and even improve the control-relevant parts.

\subsection{Behavior Cloning and Physical Robot Control}
\label{sec:behavior-cloning-robot-control}
While visual planning relies on goal images to optimize actions online, goal-agnostic tasks require executing behaviors directly. In this section, the robot must implicitly infer task progress and react to changing object poses and contacts without explicit goal conditioning. We train \method{} without goal context $\vc_t$ while retaining other objectives. At test time, the policy directly samples action chunks from observation history for execution, bypassing the test-time planning.

\phantomsection
\label{sec:task-completion-results}
\lead{Task Completion in Simulation.} We benchmark \method{} on three simulated tasks requiring complex manipulation without goal prompts: \dataset{DrawerCleanup}, \dataset{Transport}, and \dataset{ToolHang}. Figure~\ref{fig:model-ablations}(d) compares task-completion success with \methodname{Diffusion Policy}~\citep{chi2023diffusionpolicy}. \method{} achieves competitive performance on two tasks and improves success by 16 percentage points on \dataset{Transport}. These results show that \method{} supports goal-agnostic behavior cloning alongside goal-conditioned planning. Per-task standard deviations appear in Appendix Table~\ref{tab:task-completion}.

\lead{Physical-Robot Control.} Several prior JEPAs, such as \methodname{DINO-WM}~\citep{zhou2024dinowm}, \methodname{PLDM}~\citep{sobal2025pldm}, and \methodname{LeWM}~\citep{maes2026lewm}, evaluate control only in simulation, while \methodname{V-JEPA 2-AC}~\citep{assran2025vjepa2} plans on physical robots over a frozen pretrained encoder. We evaluate LeWAM on a real robotic platform through direct policy execution, testing whether an end-to-end JEPA can handle real-world contact dynamics and visual variation in closed-loop control.

We evaluate both the end-to-end LeWAM and a variant that supplies pretrained
features from a frozen \methodname{V-JEPA 2.1}~\citep{murlabadia2026vjepa21}
backbone on \dataset{Pick and Place T}, \dataset{Push-T}, and \dataset{Tape Stack}. The physical-robot
baselines are \methodname{Diffusion Policy}~\citep{chi2023diffusionpolicy} and $\pi_{0.5}$~\citep{physicalintelligence2025pi05},
a vision-language-action policy. All methods are trained or finetuned on our collected teleoperation data and directly execute the first half of each predicted
action chunk during inference. Implementation details appear in Appendix~\ref{app:real-robot}.

\begin{figure}[t]
\centering
\includegraphics[width=\linewidth]{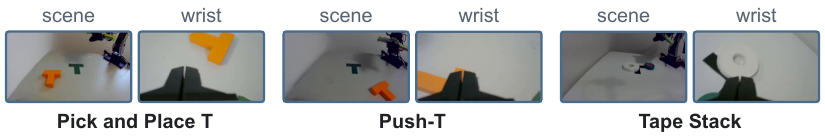}
\setlength{\abovecaptionskip}{3pt}
\caption{Paired scene and wrist view examples from three physical-robot tasks.}
\vspace{-1em}
\label{fig:real-robot-task-suite}
\end{figure}
\begin{table}[H]
\centering
\vspace{-1em}
\caption{Physical robot evaluation. Successful trials out of 30 on Pick and Place T, Push-T, and Tape Stack. GFLOPs count one inference for a full action chunk, including all denoising steps. One full LeWAM action chunk uses 270.2 GFLOPs versus 3603.6 GFLOPs for
$\pi_{0.5}$. Figure~\ref{fig:headline-results}(d) expresses
the reciprocal as action chunks per TFLOP, a $13.3\times$ difference in operation-count efficiency.}
\label{tab:real-robot-results}
\small
\setlength{\tabcolsep}{4pt}
\renewcommand{\arraystretch}{1.12}
\begin{tabular}{@{}lcccc|c@{}}
\toprule
Method & Pick and Place T & Push-T & Tape Stack & Average & GFLOPs \\
\midrule
\methodname{Diffusion Policy}~\citep{chi2023diffusionpolicy}
    & $19/30$ & $16/30$ & \textsc{0/30} & \textsc{38.9\%} & $270.0$ \\
\methodname{$\pi_{0.5}$}~\citep{physicalintelligence2025pi05}
    & $14/30$ & \textsc{3/30} & \textsc{9/30} & \textsc{28.9\%} & $3603.6$ \\
\midrule
\methodname{\method{} (Frozen V-JEPA 2.1 Encoder)}
    & $15/30$ & $12/30$ & \textsc{5/30} & \textsc{35.6\%} & $270.2$ \\
\textbf{\method{}}
    & $16/30$ & $8/30$ & \textsc{6/30} & \textsc{33.3\%} & $270.2$ \\
\bottomrule
\end{tabular}
\end{table}

\vspace{-10pt}
Table~\ref{tab:real-robot-results} shows that LeWAM is competitive with both baselines. With the same demonstrations, its encoder trained from scratch achieves a similar average success rate to the frozen V-JEPA 2.1 variant. Joint prediction of latent states and actions thus supports closed-loop physical control from teleoperation demonstrations.

On held-out episodes, both LeWAM variants reconstruct the manipulated object and coarse robot motion, with less fine detail in decoded rollouts (Appendix Figure~\ref{fig:robot-rollout-scene}).


\section{Conclusion and Limitations}
\label{sec:conclusion}

In this work we introduce \method{}, an end-to-end JEPA for joint prediction of future latent states and actions, that supports control through both planning and policy execution. On goal-reaching environments, LeWAM demonstrates improved performance in dynamics-based planning over previous JEPA-based world models. On behavioral cloning from expert demonstrations on both simulated and physical robots, LeWAM is competitive with established policy baselines. 

\Needspace{28\baselineskip}
\phantomsection
\label{sec:limitations}
\lead{Limitations and Future Work.}

\begin{itemize}[leftmargin=*,itemsep=0.25em,topsep=0.25em]
    \item \textbf{Data Coverage.} Our task suite combines expert demonstrations and non-expert interactions at limited scale. Broader play data and exploration on contact-rich tasks would extend coverage of difficult contact transitions and support training across a wider range of behaviors.

    \item \textbf{Generalization.} Learning action-conditioned dynamics could support broader generalization than policy-only approaches, motivating tests on unfamiliar objects, camera viewpoints, and contact states.

    \item \textbf{Single-Task and Single-Embodiment Learning.} Models are trained or adapted separately for each task and embodiment. Multitask training across robots and language-conditioned tasks would extend the shared dynamics and policy to broader settings.

    \item \textbf{Integration with Hierarchical Planning.} \method{} provides low-level planning and control. A high-level planner could decompose longer tasks into subgoals, each executed through LeWAM's continuous control.
    \item \textbf{Scalability of Training \& Planning.} The visual-planning model has approximately $17\text{M}$ parameters. Larger models and datasets would extend training scale, while larger candidate populations and more refinement steps would extend test-time search.
\end{itemize}

\subsection*{Reproducibility statement}
The experiments in Section~\ref{sec:experiments} use the
architectures and baseline configurations in Appendix~\ref{app:implementation},
with datasets and evaluation protocols in Appendix~\ref{app:benchmark-evaluation}.
The project website and code repository are linked on the first page.


\subsection*{Ethics statement}
The experiments collect no new human-subject data.
Reused environments, datasets, models, and software remain subject to their original
licenses.
Physical deployment (Section~\ref{sec:behavior-cloning-robot-control}) requires independent safety mechanisms.

\subsection*{AI use statement}
LLM-based assistants helped with manuscript polishing and language refinement. The authors designed and conducted the research, reviewed the analyses and manuscript, and take responsibility for its content.

\bibliography{references}
\bibliographystyle{iclr2027_conference}

\clearpage
\appendix
\phantomsection
\label{app:roadmap}

\begingroup
\centering
\vspace*{-0.6em}
{\LARGE\bfseries\sffamily Appendix\par}
\vspace{0.35em}
{\small\itshape\sffamily Supplementary material for
\textbf{LeWAM: End-to-End World and Action Modeling with JEPAs}\par}
\vspace{1.0em}
\endgroup

\noindent\begin{tcolorbox}[
    enhanced,
    colback=gray!3,
    colframe=gray!55,
    boxrule=0.4pt,
    arc=2.5pt,
    left=12pt, right=12pt, top=14pt, bottom=12pt,
    title={\textsf{\bfseries\;Contents\;}},
    coltitle=white,
    colbacktitle=gray!55,
    fonttitle=\bfseries\sffamily\small,
    attach boxed title to top center={yshift=-3pt},
    boxed title style={size=small, sharp corners, boxrule=0pt, arc=2pt},
]
\renewcommand{\arraystretch}{1.2}
\sffamily\small
\begin{tabular*}{\linewidth}{@{}l@{\hspace{6pt}}l@{\extracolsep{\fill}}r@{}}
\appendixbadge{A} & \hyperref[app:implementation]{\textbf{Implementation Details}}
& p.~\pageref{app:implementation} \\
& {\scriptsize\color{gray!35!black}\hspace{1em}\hyperref[app:planners]{LeWAM Planning}}
& {\scriptsize\color{gray!35!black}p.~\pageref{app:planners}} \\
& {\scriptsize\color{gray!35!black}\hspace{1em}\hyperref[app:baseline-implementations]{Baseline Implementations}}
& {\scriptsize\color{gray!35!black}p.~\pageref{app:baseline-implementations}} \\[0.3em]
\appendixbadge{B} & \hyperref[app:benchmark-evaluation]{\textbf{Benchmarks and Evaluation}}
& p.~\pageref{app:benchmark-evaluation} \\
& {\scriptsize\color{gray!35!black}\hspace{1em}\hyperref[app:tasks-data]{Tasks and Datasets}}
& {\scriptsize\color{gray!35!black}p.~\pageref{app:tasks-data}} \\
& {\scriptsize\color{gray!35!black}\hspace{1em}\hyperref[app:splits]{Goals and Success Criteria}}
& {\scriptsize\color{gray!35!black}p.~\pageref{app:splits}} \\
& {\scriptsize\color{gray!35!black}\hspace{1em}\hyperref[app:runtime]{Planning and Timing Protocols}}
& {\scriptsize\color{gray!35!black}p.~\pageref{app:runtime}} \\[0.3em]
\appendixbadge{C} & \hyperref[app:numerical-results]{\textbf{Additional Results}}
& p.~\pageref{app:numerical-results} \\
& {\scriptsize\color{gray!35!black}\hspace{1em}\hyperref[app:visual-planning-results]{Visual Planning Results}}
& {\scriptsize\color{gray!35!black}p.~\pageref{app:visual-planning-results}} \\
& {\scriptsize\color{gray!35!black}\hspace{1em}\hyperref[app:component-ablations]{Component and Visual-Encoder Ablations}}
& {\scriptsize\color{gray!35!black}p.~\pageref{app:component-ablations}} \\
& {\scriptsize\color{gray!35!black}\hspace{1em}\hyperref[app:latent-probing]{Probing Latent Representations}}
& {\scriptsize\color{gray!35!black}p.~\pageref{app:latent-probing}} \\
& {\scriptsize\color{gray!35!black}\hspace{1em}\hyperref[app:simulation-rollouts]{Visual Reconstructions and Open-Loop Rollouts}}
& {\scriptsize\color{gray!35!black}p.~\pageref{app:simulation-rollouts}} \\
& {\scriptsize\color{gray!35!black}\hspace{1em}\hyperref[app:action-distributions]{Multimodal Action Distributions}}
& {\scriptsize\color{gray!35!black}p.~\pageref{app:action-distributions}} \\
& {\scriptsize\color{gray!35!black}\hspace{1em}\hyperref[app:simulation-completion-results]{Simulation Task-Completion Results}}
& {\scriptsize\color{gray!35!black}p.~\pageref{app:simulation-completion-results}} \\[0.3em]
\appendixbadge{D} & \hyperref[app:real-robot]{\textbf{Physical-Robot Setup and Evaluation}}
& p.~\pageref{app:real-robot} \\
& {\scriptsize\color{gray!35!black}\hspace{1em}\hyperref[app:robot-data]{Platform and Data Collection}}
& {\scriptsize\color{gray!35!black}p.~\pageref{app:robot-data}} \\
& {\scriptsize\color{gray!35!black}\hspace{1em}\hyperref[app:robot-implementation]{Model Implementation}}
& {\scriptsize\color{gray!35!black}p.~\pageref{app:robot-implementation}} \\
& {\scriptsize\color{gray!35!black}\hspace{1em}\hyperref[app:bc-physical]{Physical-Robot Baseline Configurations}}
& {\scriptsize\color{gray!35!black}p.~\pageref{app:bc-physical}} \\
& {\scriptsize\color{gray!35!black}\hspace{1em}\hyperref[app:robot-evaluation]{Closed-Loop Evaluation}}
& {\scriptsize\color{gray!35!black}p.~\pageref{app:robot-evaluation}} \\
& {\scriptsize\color{gray!35!black}\hspace{1em}\hyperref[app:robot-diagnostics]{Representation Diagnostics and Evaluation}}
& {\scriptsize\color{gray!35!black}p.~\pageref{app:robot-diagnostics}} \\[0.3em]
\appendixbadge{E} & \hyperref[app:joint-prediction]{\textbf{Discussion of the Role of Joint Prediction}}
& p.~\pageref{app:joint-prediction} \\
& {\scriptsize\color{gray!35!black}\hspace{1em}\hyperref[app:joint-representation]{Representation and Dynamics}}
& {\scriptsize\color{gray!35!black}p.~\pageref{app:joint-representation}} \\
& {\scriptsize\color{gray!35!black}\hspace{1em}\hyperref[app:joint-planning]{Planning with Learned Action Proposals}}
& {\scriptsize\color{gray!35!black}p.~\pageref{app:joint-planning}} \\[0.3em]
\appendixbadge{F} & \hyperref[app:author-contributions]{\textbf{Author Contribution Statement}}
& p.~\pageref{app:author-contributions} \\
\end{tabular*}
\end{tcolorbox}

\newpage

\section{Implementation Details}
\label{app:implementation}

\phantomsection
\label{app:arch}

\lead{Temporal Sampling.} Given observation-history stride $\Delta$, the predictor in
Section~\ref{sec:joint-predictor} uses two history latents
$\vz_{\leq t}=(\vz_{t-\Delta},\vz_t)$, a dynamics stride of $K=5$ raw steps, and an action
chunk of $C=2K=10$ raw steps. We use a stride of $\Delta=K$ for goal-reaching tasks and $\Delta=1$ for task-completion frames. To predict $\vz_{t+5}$, the dynamics branch receives all five actions $\va_t,\ldots,\va_{t+4}$, and the flow policy models
all ten actions $\va_t,\ldots,\va_{t+9}$. Only frames are subsampled. During imagined
rollouts, the history update retains the two most recent latent states.

\lead{Inner Architecture.} The observation encoder is
a ResNet-18-style network~\citep{he2016resnet} trained from scratch. Its convolutional backbone
uses GroupNorm with 16 groups. Input images are cropped from $224\times224$ to $202\times202$,
with random offsets during training and a center crop at evaluation. A $1\times1$ convolution
produces 32 spatial maps, and a spatial softmax pooling converts these to 32 expected normalized image-coordinate
pairs. After that, a two-layer projection MLP with BatchNorm maps these 64 features through a hidden width
of 384 to a 192-dimensional latent state, which is the same as LeWM~\citep{maes2026lewm} in simulations for a fair comparison. The MoT predictor has width 192, 4 blocks, 4 attention heads, and dropout 0.1. Its token sequence contains two history
latents, one learned state query, five clean action tokens, and ten action flow slots. Learned
type and positional embeddings distinguish history, state query, clean action, and action flow
tokens. The state readout uses LayerNorm followed by a linear map to the 192-dimensional latent,
while the action readout maps each action flow slot to a velocity vector in raw-action space.

Goal information $\vc_t$ is optional and is omitted for goal-free task completion. For goal-conditioned visual planning, $\vc_t$ contains the goal image latent $\vz_g$ and the normalized remaining horizon. The goal latent is supplied as an additional token in the action branch, while only noisy action tokens can attend to it. The horizon $h$ between the current timestep and the goal is normalized into a $[0,1]$ range relative to a maximal value $H_{\mathrm{max}}$ which dictates the longest distance-from-goal seen during training. $h$ modulates the noisy action tokens using AdaLN-Zero~\citep{peebles2023dit} blocks with hidden width 512 and a 128-dimensional sinusoidal embedding.

In total, the model has approximately 16.9M trainable parameters in visual planning. For the simulated experiments, action sampling uses 8 Euler steps to balance speed and quality.
The reported MSE dynamics configuration uses an online encoder target, no state noise, no state
flow integration, no momentum target encoder, only SIGReg with 1,024 projections and 17 knots. For the training objective in Equation~\ref{eq:joint-objective}, we set $\lambda_{\mathrm{sig}} = 0.04$.

\lead{Training Recipe in Simulated Environments.} For goal-conditioned visual planning, LeWAM runs use a batch size of 128 and train for 50 epochs, with a base learning rate of $1.5\times10^{-4}$ and five warmup epochs. The goal-free task-completion runs use a batch size of 64 and train for 120 epochs, with a base learning rate of $10^{-4}$ and ten warmup epochs. Both use AdamW with weight decay $10^{-4}$ and a uniform learning rate across the encoder and MoT predictor. The learning rate increases linearly during warmup and then follows an epochwise cosine schedule. Training uses full precision, a gradient-norm clip of 1, and seed 42. We split valid decision points into 90\% training and 10\% validation, then sample training points with replacement within each epoch. After each epoch, we retain the checkpoint with the lowest validation action flow-matching loss, where task success is not used for checkpoint selection.

\lead{Tokens and Conditioning in MoT.} With the optional goal context $\vc_t$ omitted, the MoT sequence is
\begin{equation}
    \left[
    \vz_{t-\Delta},\vz_t,\vq^z_{t+K},
    \mA_{t,K},\mA_{t,C}^{(s)}
    \right].
    \label{eq:mot-tokens}
\end{equation}
Here $\mA_{t,K}$ and $\mA_{t,C}^{(s)}$ expand into $K$ clean and $2K$ noisy action tokens, respectively.
The first three tokens are processed by $\statebranch$, and both action groups by $\actionbranch$. Each MoT branch has separate query, key, value, output, normalization, and
feedforward parameters, while one global masked self-attention operates over the complete token sequence. Flow timestep
$s$ modulates only the noisy action slots through AdaLN-Zero.

\begin{figure}[!t]
\centering
\includegraphics[width=0.90\linewidth]{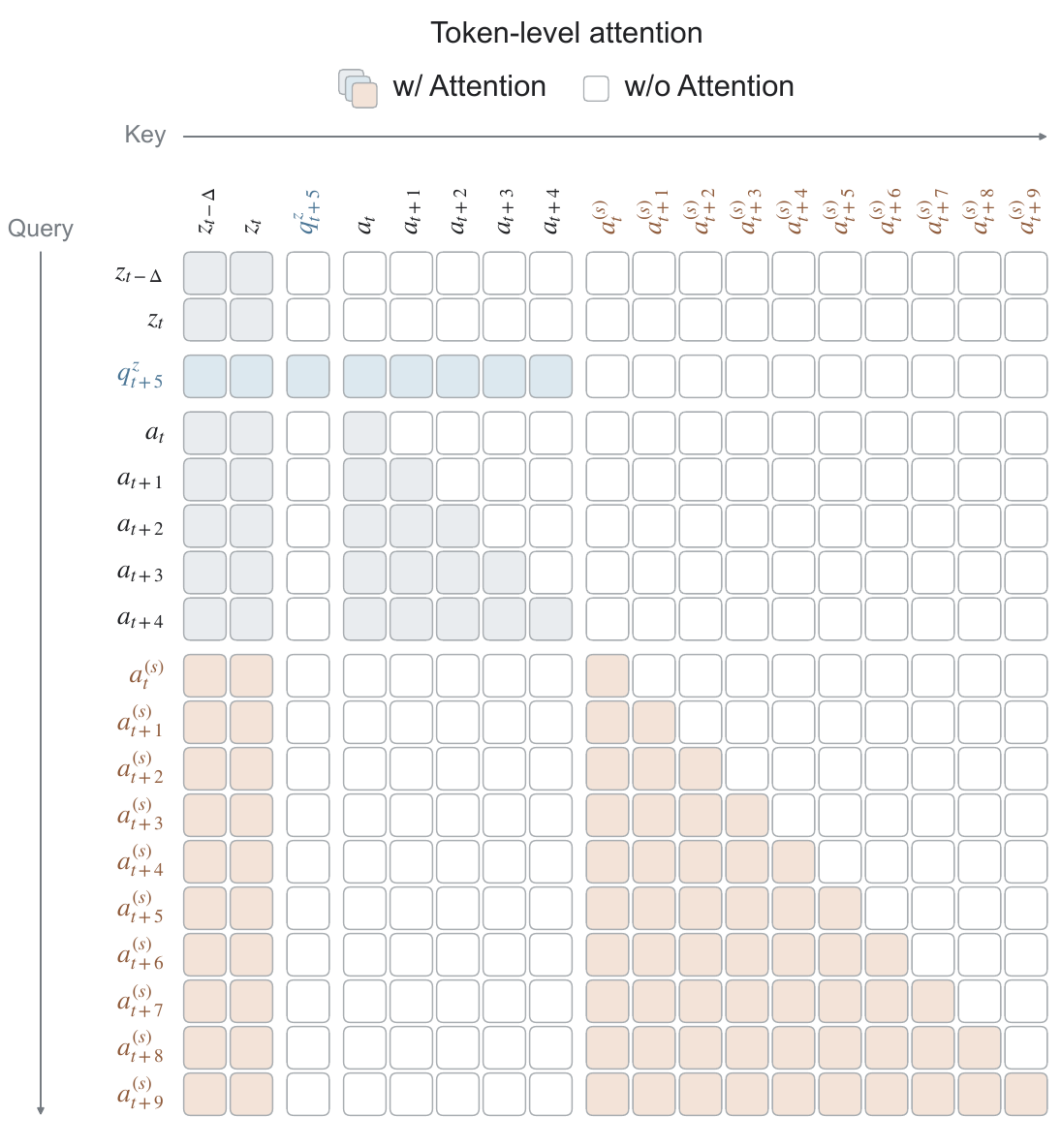}
\caption{\lead{Token-Level MoT Attention.} Expansion of Figure~\ref{fig:mot-architecture}
for two history latents, one learned state query, five clean actions, and ten interpolated
action tokens. Rows are queries; columns are keys. Filled cells permit attention; white cells are masked. Blue and brown mark
latent state and action velocity prediction queries; gray marks context. Wider gaps
separate token groups. Optional goal conditioning $\vc_t$ is omitted.
$\Delta$ is the history stride, and action indices remain at the raw environment
rate.}
\label{fig:mot-mask-full}
\end{figure}

\Needspace{9\baselineskip}
\lead{Attention Mask.} Let $\mathcal{V}(x)$ denote the tokens visible to query $x$. The mask below describes the sequence without optional goal context $\vc_t$. When this context is provided, the goal token is additionally visible to the noisy action tokens only.
Action prefixes begin at the current chunk's start $t$; padded history keys are excluded:
\begin{equation}
\begin{aligned}
\mathcal{V}(\vz_i)
    &=\{\vz_{t-\Delta},\vz_t\},
    && i\in\{t-\Delta,t\},\\
\mathcal{V}(\vq^z_{t+K})
    &=\{\vz_{t-\Delta},\vz_t,\vq^z_{t+K},\mA_{t,K}\},\\
\mathcal{V}(\va_j)
    &=\{\vz_{t-\Delta},\vz_t,\va_{\leq j}\},
    &&j\in\{t,\ldots,t+K-1\},\\
\mathcal{V}(\va_j^{(s)})
    &=\{\vz_{t-\Delta},\vz_t,\va_{\leq j}^{(s)}\},
    &&j\in\{t,\ldots,t+2K-1\}.
\end{aligned}
\label{eq:mot-mask}
\end{equation}
Clean action tokens therefore cannot relay target information to history or action flow tokens across layers. A visualization of the attention mask is provided in Figure~\ref{fig:mot-mask-full}.

\clearpage
\subsection{LeWAM Planning}
\label{app:planners}

\lead{Optimization Details.} The goal cost $\gC$ is defined in
Eq.~(\ref{eq:lewam-gradient-planning}). During refinement, model parameters
remain frozen and only the normalized action plans receive gradients. For each
sampled plan $\mA^{\mathrm{sample}}$, Adam minimizes the goal cost with a
quadratic proximity penalty:
\begin{equation}
    \gJ_{\mathrm{prox}}(\mA;\mA^{\mathrm{sample}})
    =\gC(\mA)+\lambda_{\mathrm{prox}}
    \left\|\mA-\mA^{\mathrm{sample}}\right\|_F^2.
    \label{eq:proximity-penalty-planning}
\end{equation}
The penalty affects the updates, while candidate retention and final selection
use the unpenalized goal cost $\gC$, and we clip normalized actions to
$[-a_{\max},a_{\max}]$ after each update. This penalty aims at discouraging the huge changes on actions, since the action flow already provide an effective prior.

\begin{algorithm}[H]
\caption{Candidate refinement and selection for LeWAM planning}
\label{alg:lewam-gradient-planning}
\begin{algorithmic}[1]
\Require $\vz_{\leq t},\vc_t,\vz_g,L,N,J,\eta,\lambda_{\mathrm{prox}},a_{\max}$
\Statex \textbf{Sample candidates from the learned policy}
\State $\mA_{1:N}^{\mathrm{sample}} \gets N$ policy rollouts from $(\vz_{\leq t},\vc_t)$ for $L/K$ blocks
\State $c_{1:N} \gets \gC(\mA_{1:N}^{\mathrm{sample}})$
\State $\mA^\star_{1:N} \gets \mA_{1:N}^{\mathrm{sample}}$, \quad $\mA_{1:N} \gets \mA_{1:N}^{\mathrm{sample}}$
\State $c^\star_{1:N} \gets c_{1:N}$

\Statex \textbf{Refine each candidate}
\For{$j = 0,\ldots,J-1$}
    \State $\mA_{1:N} \gets
    \operatorname{AdamStep}\left(
    \mA_{1:N},
    \nabla_{\mA_{1:N}}\gJ_{\mathrm{prox}}
    (\mA_{1:N};\mA_{1:N}^{\mathrm{sample}}),
    \eta\right)$
    \State $\mA_{1:N} \gets
    \operatorname{clip}(\mA_{1:N},-a_{\max},a_{\max})$
    \State $c_{1:N} \gets \gC(\mA_{1:N})$
    \State $\mA^\star_n \gets \mA_n,\quad
    c_n^\star \gets c_n
    \quad \forall n\text{ such that }c_n<c_n^\star$
\EndFor

\Statex \textbf{Select the lowest-cost iterate}
\State $n^\star \gets \argmin_{n=1,\ldots,N} c_n^\star$
\State \Return $\mA^\star_{n^\star}$
\end{algorithmic}
\end{algorithm}

We let the Adam optimizer use a learning rate of $0.05$, gradient-norm
clipping at $10$, $a_{\max}=3$, and $\lambda_{\mathrm{prox}}=0.01$.

\subsection{Baseline Implementations}
\label{app:baseline-implementations}

The baselines in Section~\ref{sec:worldforge8} use the released \methodname{stable-worldmodel}
codebase~\citep{maes2026swm}, retaining its environment wrappers, dataset interface,
action bounds, and evaluation loop. Unless stated below, each method follows its official
implementation and hyperparameters.\footnote{Official repositories:
\href{https://github.com/gaoyuezhou/dino_wm}{DINO-WM},
\href{https://github.com/galilai-group/stable-worldmodel}{stable-worldmodel}, and
\href{https://github.com/hdnndh/Latent-Geometry-Beyond-Search-Amortizing-Planning-in-World-Models}{GC-IDM}.}
These configurations describe goal-conditioned comparisons. Methods share observations, actions,
start and goal pairs, and success predicates. We list deviations from released configurations.

\baselineitem{MethodGCIDM}{\textbf{\methodname{LeWM (IDM)}.} This is the
\methodname{GC-IDM}~\citep{nguyen2026gcidm} configuration: a pretrained
\methodname{LeWM} checkpoint trained by us is frozen, then a goal-conditioned inverse-dynamics model is post-trained.
We follow its official two-stage implementation.
Goal reaching samples a remaining horizon uniformly
from 1 to 50 and clips it to the episode boundary.}

\baselineitem{MethodLeWM}{\textbf{\methodname{LeWM (CEM)}.} We follow the official
\methodname{LeWM} training and CEM evaluation settings~\citep{maes2026lewm}. We train a checkpoint for each task and use these checkpoints in all quantitative
experiments with \methodname{LeWM} as a baseline. }

\baselineitem{MethodDINOWM}{\textbf{\methodname{DINO-WM}.} We evaluate the released patch-token
model and a matched CLS-token variant; tokenization is their only architectural difference. The
six added tasks use ten training epochs, while the original four retain their four-epoch
checkpoints.}

\baselineitem{MethodGCBC}{\textbf{\methodname{GCBC}.} We use the goal-conditioned
behavior-cloning baseline from the released codebase~\citep{maes2026swm}, with its pretrained
\methodname{DINOv2} encoder frozen and the policy trained for 20 rather than 100 epochs.
Short-horizon goal reaching samples a uniformly later frame from the same episode.
At evaluation, the policy executes its predicted five-step action chunks without planning.}

\Needspace{5\baselineskip}
\lead{Behavior-Cloning Baseline in Simulation.}
\label{app:bc-simulation}
The task-completion comparison in Table~\ref{tab:task-completion} uses a policy
trained on demonstrations from each task, conditioned on observation history
without a goal image or goal-horizon input.

\baselineitem{MethodDiffusionPolicy}{\textbf{\methodname{Diffusion Policy}}~\citep{chi2023diffusionpolicy}.
We use the convolutional variant (DP-C), with a ResNet-18 encoder, GroupNorm, and
spatial softmax for each camera. The image features condition a one-dimensional
U-Net with channel widths $(512,1024,2048)$, kernel size five, and diffusion-step
embedding dimension 128. Training minimizes noise-prediction MSE with a
100-step squared-cosine diffusion schedule; inference uses 100 DDPM steps.
The policy takes two consecutive observations, predicts a 16-action sequence,
and executes eight actions starting at the current timestep before resampling.}

\dataset{ToolHang} uses the scene camera; \dataset{DrawerCleanup} and
\dataset{Transport} use the scene camera and both wrist cameras. Inputs are
$224\times224$ RGB images, with random $202\times202$ crops during training and
center crops at evaluation, normalized to $[-1,1]$. The policy uses visual
observations only. Training windows advance at the raw frame rate, with boundary
frames and actions repeated for padding. We randomly split windows into 90\%
training and 10\% validation using seed 42. Each policy trains for 120 epochs
with batch size 64 and AdamW, learning rate $10^{-4}$, betas $(0.95,0.999)$, and
weight decay $10^{-6}$. The learning rate follows cosine decay after 500 warmup
updates. Actions retain their recorded scale for \dataset{ToolHang} and
\dataset{Transport}; \dataset{DrawerCleanup} uses min-max scaling to $[-1,1]$.
We select the checkpoint with the lowest validation denoising loss and evaluate
its raw weights without exponential moving averaging. Each checkpoint is
evaluated with seeds 0, 1, and 42, using 50 episodes per seed.

\section{Benchmarks and Evaluation}
\label{app:benchmark-evaluation}
\label{app:datasets}
\label{app:evaluation}

\subsection{Tasks and Datasets}
\label{app:tasks-data}
\label{app:lewm-bench}
\label{app:wf-data}

\lead{Task Selection.} We retain the four environments used by
\methodname{LeWM}~\citep{maes2026lewm} for direct comparison, then add six tasks
that broaden navigation and manipulation to larger mazes, interacting objects,
coupled button states, precise tool placement, bimanual transfer, and articulated
drawers (Figure~\ref{fig:worldforge8}). The ten tasks also span different action
dimensions, observation views, trajectory lengths, and data-collection regimes.
Together, they test the same model design under varied control and data conditions.

\lead{Data Provenance.}
We describe each dataset in the environment order of Figure~\ref{fig:worldforge8}.
\begin{itemize}
    \item \dataset{Two-Room}. We use the dataset released with
    \methodname{LeWM}~\citep{maes2026lewm}, collected by a noisy heuristic
    navigation policy.
    \item \dataset{Reacher}. We use the dataset released with
    \methodname{LeWM}~\citep{maes2026lewm}, collected by a random policy with uniform distribution.
    \item \dataset{Push-T}. We use the non-expert interaction dataset released by
    \methodname{DINO-WM}~\citep{zhou2024dinowm}, generated by replaying the original
    expert trajectories with different levels of action noise.
    \item \dataset{Cube}. We use the dataset released with
    \methodname{LeWM}~\citep{maes2026lewm} for the single-cube
    \dataset{OGBench}~\citep{park2025ogbench} environment, collected by a scripted
    expert that picks and places the cube at sampled target locations. We just change the success criteria.
    \item \dataset{PointMaze-Large}. We use the official
   \dataset{OGBench}~\citep{park2025ogbench} \texttt{pointmaze-large-navigate-v0}
    dataset, collected by a noisy expert that navigates between randomly sampled
    locations. 
    \needspace{5\baselineskip}
    \item \dataset{Scene}. We collect data ourselves using the
    \methodname{stable-worldmodel}~\citep{maes2026swm} wrapper for
    \dataset{OGBench}~\citep{park2025ogbench} \dataset{Scene}
    (\texttt{swm/OGBScene-v0}) and its \texttt{ExpertPolicy}, a noisy, closed-loop
    scripted oracle. The policy interacts with the cube, drawer, window, and
    buttons; we record a single camera view.
    \item \dataset{Puzzle}. We collect data ourselves using the
    \methodname{stable-worldmodel}~\citep{maes2026swm} wrapper for the
    $3\times3$ \dataset{OGBench}~\citep{park2025ogbench} \dataset{Puzzle}
    environment and its scripted \texttt{ExpertPolicy}. The policy presses
    buttons in sampled sequences; we record a single camera view.
    \item \dataset{ToolHang}. We use the official human behavior-cloning
    demonstrations from \dataset{robomimic}~\citep{mandlekar2022robomimic}.
    \item \dataset{Transport}. We use the official generated behavior-cloning
    demonstrations from \dataset{DexMimicGen}~\citep{jiang2025dexmimicgen}.
    \item \dataset{DrawerCleanup}. We use the official generated behavior-cloning
    demonstrations from \dataset{DexMimicGen}~\citep{jiang2025dexmimicgen}.
\end{itemize}

The three behavior-cloning datasets broaden the visual-planning comparison to
complex contact-rich behaviors with higher-dimensional actions, although their
task-focused demonstrations may limit the diversity of sampled visual goals.

Table~\ref{tab:dataset-scale} summarizes all ten tasks, including observation
views, raw action dimensions, and dataset scale.
Recorded lengths describe the data, while Appendix~\ref{app:splits} specifies
evaluation horizons and execution budgets. Recorded duration uses the native
sampling rate before temporal subsampling; synchronized views of the same
trajectories are counted once. \dataset{ToolHang} uses one oblique scene view
(about $45^\circ$), while \dataset{Transport} and \dataset{DrawerCleanup} use a
scene view and both wrist views.

\begin{table}[H]
\centering
\caption{\lead{Simulation Benchmark Datasets.} Ten tasks in
Figure~\ref{fig:worldforge8} order. Views gives camera-stream count; Dim. is the
raw action dimension. Statistics cover complete source datasets before splits.
Total hours are computed before rounding per-task values.}
\label{tab:dataset-scale}
\small
\setlength{\tabcolsep}{2pt}
\renewcommand{\arraystretch}{1.10}
\begin{tabular*}{\linewidth}{@{\extracolsep{\fill}}llccrrrrr@{}}
\toprule
Task & Collection regime & Views & Dim. & Traj. & \shortstack{Mean\\frames} & Total frames & Hz & Hours \\
\midrule
\dataset{Two-Room} & Noisy heuristic & 1 & 2 & 10{,}000 & 92 & 920{,}809 & 10 & 25.58 \\
\dataset{Reacher} & Uniform random & 1 & 2 & 10{,}000 & 201 & 2{,}010{,}000 & 25 & 22.33 \\
\dataset{Push-T} & Non-expert & 1 & 2 & 18{,}685 & 125 & 2{,}336{,}736 & 10 & 64.91 \\
\dataset{Cube} & Scripted expert & 1 & 5 & 10{,}000 & 201 & 2{,}010{,}000 & 20 & 27.92 \\
\dataset{PointMaze-Large} & Scripted navigation & 1 & 2 & 1{,}000 & 1{,}000 & 1{,}000{,}000 & 10 & 27.78 \\
\dataset{Scene} & Scripted expert & 1 & 5 & 10{,}024 & 200 & 2{,}004{,}800 & 20 & 27.84 \\
\dataset{Puzzle} (3$\times$3) & Scripted expert & 1 & 5 & 10{,}024 & 200 & 2{,}004{,}800 & 20 & 27.84 \\
\dataset{ToolHang} & Human demos & 1 & 7 & 200 & 480 & 95{,}962 & 20 & 1.33 \\
\dataset{Transport} & Generated demos & 3 & 14 & 1{,}029 & 408 & 419{,}712 & 20 & 5.83 \\
\dataset{DrawerCleanup} & Generated demos & 3 & 24 & 1{,}026 & 291 & 298{,}235 & 20 & 4.14 \\
\midrule
Reported total (10) & & & & 71{,}988 & & 13{,}101{,}054 & & 235.51 \\
\bottomrule
\end{tabular*}
\end{table}

\subsection{Goals and Success Criteria}
\label{app:splits}

For the comparisons in Section~\ref{sec:worldforge8}, unless stated otherwise, if not specify, 
we report mean success and standard deviation over three seeds with 50 episodes per seed.
Methods within a protocol share observations, action limits, execution budgets, and
success predicates.

\lead{Goal-Conditioned Visual Planning.} In Section~\ref{sec:goal-conditioned-results},
the initial state is sampled from an offline trajectory. Except on \dataset{Two-Room}, the frame 25 steps later
supplies the visual goal, and each controller receives 50 environment actions. On \dataset{Two-Room}, we use a goal offset of 100 environment steps and an execution budget of 150 actions, following \methodname{LeWorldModel}~\citep{maes2026lewm}.
We report first-hit success for comparison with the released
\methodname{LeWM} evaluation~\citep{maes2026lewm}; the \dataset{Cube} predicate correction
is detailed below. \dataset{Scene} and \dataset{Puzzle} follow the same protocol;
their random-policy and zero-action controls appear alongside those for the other
environments in Table~\ref{tab:policy-floor-controls}.

\lead{Policy-Floor Controls.} To ensure the benchmarks are reliable in reflecting the model ability, Table~\ref{tab:policy-floor-controls} also reports
zero-action and random-policy success under the full-trajectory task-completion
evaluation. These blind controls cover the benchmark environments.

\begin{figure}[H]
\centering
\includegraphics[width=0.72\linewidth]{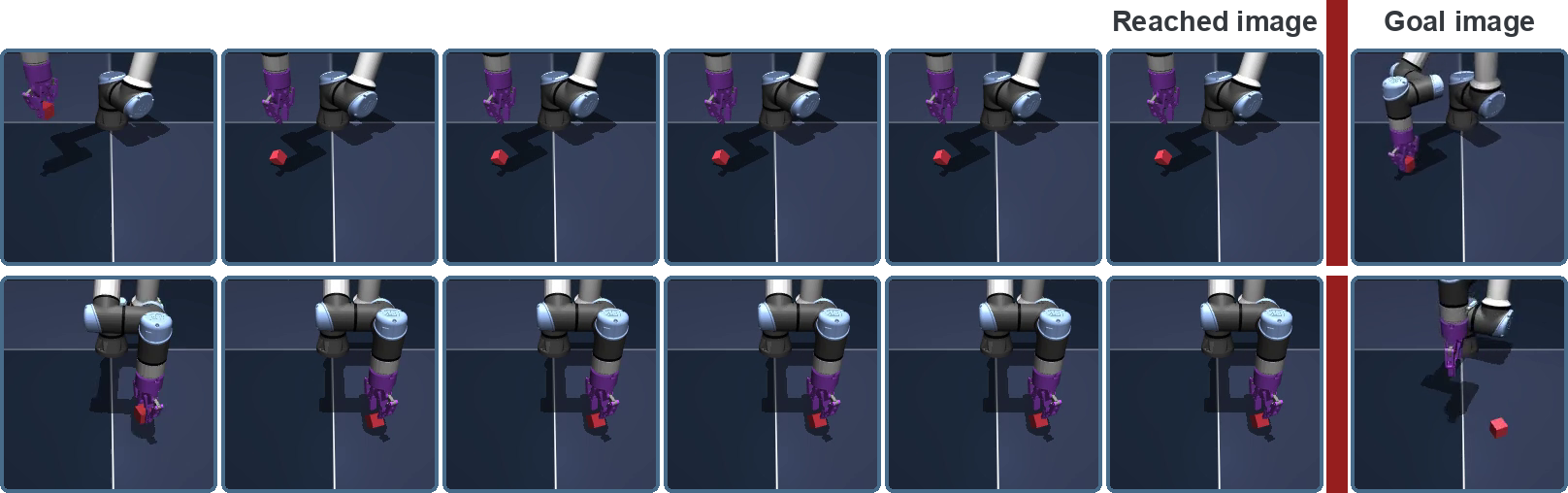}
\caption{\lead{A Degenerate \dataset{Cube} Success under Goal Reaching.}
Each row shows a policy rollout (left, frames over time) beside its goal frame (far right,
after the divider). Throughout the
execution, the gripper never reaches the goal end-effector pose, yet the cube stays within the
$0.04$~m object-only threshold of the goal, so object-position-only predicate scores
the episode a spurious success. Our end-effector term (Table~\ref{tab:policy-floor-controls}) rejects these
degenerate successes by additionally requiring the gripper to reach the goal pose.}
\label{fig:cube-object-only}
\end{figure}

\lead{Cube Success Predicate.} Specifically, we add an end-effector (EEF) check to the \dataset{Cube} success criterion to exclude spurious successes. The original \dataset{Cube} criterion checks whether the
cube center lies within $0.04$ m of the target, regardless of the gripper pose.
For 25-step goals, this credits zero action on $42.7\%$ of episodes
(Table~\ref{tab:policy-floor-controls}). The corrected criterion additionally requires
the end effector to lie within $0.04$ m of its goal-frame position, reducing this floor
to $2.7\%$. Figure~\ref{fig:cube-object-only} illustrates the object-only failure case.

\begin{table}[t]
    \centering
\begingroup
\setlength{\tabcolsep}{4pt}
\setlength{\arrayrulewidth}{0.3pt}
\renewcommand{\arraystretch}{1.35}
\newcommand{\policyfloorcell}[3][black]{%
  \pgfmathsetmacro{\policyfloortint}{min(100,100*#2/45)}%
  \edef\policyfloorfill{\noexpand\cellcolor{PaperBlueFill!\policyfloortint!white}}%
  \policyfloorfill
  \textcolor{#1}{\ensuremath{#2\pm#3}}%
}
\begin{tabular}{%
  >{\raggedright\arraybackslash}m{\dimexpr.23\linewidth-2\tabcolsep-0.9pt\relax}|
  >{\centering\arraybackslash}m{\dimexpr.15\linewidth-2\tabcolsep\relax}|
  *{2}{>{\centering\arraybackslash}m{\dimexpr.155\linewidth-2\tabcolsep\relax}}|
  *{2}{>{\centering\arraybackslash}m{\dimexpr.155\linewidth-2\tabcolsep\relax}}}
\hline
\multirow{2}{*}{Environment} & \multirow{2}{*}{\shortstack{Raw action\\dim.}}
  & \multicolumn{2}{c|}{\rule{0pt}{23pt}\shortstack{25-step goal\\reaching}}
  & \multicolumn{2}{c}{\rule{0pt}{23pt}\shortstack{Task\\Completion}} \\
\cline{3-6}
  & & Zero-action & Random & Zero-action & Random \\
\hline
Two-Room & 2 & \policyfloorcell{7.3}{0.9} & \policyfloorcell{25.3}{5.0} & \policyfloorcell{0.0}{0.0} & \policyfloorcell{2.7}{2.5} \\\hline
Reacher & 2 & \policyfloorcell{3.3}{1.9} & \policyfloorcell{12.0}{1.6} & \policyfloorcell{0.0}{0.0} & \policyfloorcell{12.0}{4.9} \\\hline
Push-T & 2 & \policyfloorcell{0.0}{0.0} & \policyfloorcell{1.3}{0.9} & \policyfloorcell{0.0}{0.0} & \policyfloorcell{0.0}{0.0} \\\hline
\textcolor{PaperGrayDark}{Cube} & \textcolor{PaperGrayDark}{5} & \policyfloorcell[PaperGrayDark]{42.7}{6.6} & \policyfloorcell[PaperGrayDark]{44.0}{1.6} & \policyfloorcell[PaperGrayDark]{4.7}{2.5} & \policyfloorcell[PaperGrayDark]{6.0}{1.6} \\\hline
Cube ($+$EEF) & 5 & \policyfloorcell{2.7}{0.9} & \policyfloorcell{4.7}{0.9} & \policyfloorcell{4.7}{2.5} & \policyfloorcell{6.0}{1.6} \\\hline
PointMaze-Large & 2 & \policyfloorcell{6.0}{0.0} & \policyfloorcell{9.3}{1.9} & \policyfloorcell{0.7}{0.9} & \policyfloorcell{6.0}{3.3} \\\hline
Scene & 5 & \policyfloorcell{2.7}{2.3} & \policyfloorcell{2.7}{2.3} & \policyfloorcell{0.0}{0.0} & \policyfloorcell{0.0}{0.0} \\\hline
Puzzle & 5 & \policyfloorcell{0.7}{1.2} & \policyfloorcell{0.7}{1.2} & \policyfloorcell{0.0}{0.0} & \policyfloorcell{0.0}{0.0} \\\hline
ToolHang & 7 & \policyfloorcell{22.7}{2.5} & \policyfloorcell{26.0}{1.6} & \policyfloorcell{0.0}{0.0} & \policyfloorcell{0.0}{0.0} \\\hline
Transport & 14 & \policyfloorcell{3.3}{0.9} & \policyfloorcell{2.7}{0.9} & \policyfloorcell{0.0}{0.0} & \policyfloorcell{0.0}{0.0} \\\hline
DrawerCleanup & 24 & \policyfloorcell{0.0}{0.0} & \policyfloorcell{6.0}{0.0} & \policyfloorcell{0.0}{0.0} & \policyfloorcell{0.0}{0.0} \\
\hline
\end{tabular}
\par\vspace{3pt}
{\footnotesize\color{PaperGrayDark}Darker shading indicates a higher control success rate.\par}
\endgroup

    \caption{\lead{Action Dimensions and Policy-Floor Controls.}
    Rows cover all ten environments in Figure~\ref{fig:worldforge8}, plus a
    \dataset{Cube} ($+$EEF) row isolating the corrected end-effector predicate.
    Raw action dimensionality is reported beside zero-action and random-policy success rates (\%)
    for short-horizon goal reaching and full-trajectory task completion, averaged over three seeds.
    These blind controls measure baseline success from zero actions and random actions.
    \dataset{Cube} has the same task-completion values in both rows because the
    end-effector correction applies only to short-horizon goal reaching.
    \dataset{DrawerCleanup} exposes 24 raw coordinates; the policy predicts 22 active dimensions
    after removing two constant joints. }
    \label{tab:policy-floor-controls}
\end{table}

\FloatBarrier
\Needspace{20\baselineskip}
\subsection{Planning and Timing Protocols}
\label{app:runtime}

\lead{Goal-Horizon Sweep.} The sweep in Figure~\ref{fig:planning-horizon} and
Table~\ref{tab:long-horizon-planning} starts from the first frame of each held-out
trajectory and uses the recorded frames 25, 50, 75, or 100 steps later as goals.
All tasks use the shared planning and replanning settings in Section~\ref{sec:control-planning}.

\lead{Planner-Sensitivity Protocol.} On \dataset{Push-T}, we vary Adam refinement updates while fixing the candidate count at 32, then vary the number of sampled candidates while fixing refinement at five updates. Goal-reaching success and per-plan latency are measured over three seeds with 50 episodes per seed. The measured curves appear in Figure~\ref{fig:pusht-grad-tradeoff} in Appendix~\ref{app:visual-planning-results}.

\lead{Timing Scope.} Table~\ref{tab:planning-speed} in Appendix~\ref{app:visual-planning-results} reports time per planner call. Figure~\ref{fig:headline-results}(c)
averages the per-task throughput across its four tasks, using
$1000/\text{milliseconds per plan}$ to obtain plans/s. The compared settings are
$300\times30$ CEM for \methodname{LeWM}~\citep{maes2026lewm} and five gradient
steps for \method{}. All compared neural networks use the same H100 GPU.

\FloatBarrier
\Needspace{20\baselineskip}
\section{Additional Results}
\label{app:ablations}
\label{app:numerical-results}
\label{app:additional-results}

\subsection{Visual Planning Results}
\label{app:visual-planning-results}

Table~\ref{tab:long-horizon-planning} gives the measured goal-horizon sweeps behind Figure~\ref{fig:planning-horizon}. Table~\ref{tab:planning-speed} gives the per-call timing behind Figure~\ref{fig:headline-results}(c). Appendix~\ref{app:runtime} specifies the evaluation and timing protocols.

\begin{table}[H]
\centering
\caption{\lead{Goal-Reaching Success across Horizons.} Success rate (\%), mean $\pm$ standard
deviation. Measured values for Figure~\ref{fig:planning-horizon}. The \dataset{Cube}
horizon sweep is distinct from the sampled-start 25-step evaluation in
Figure~\ref{fig:short-horizon-control}.}
\label{tab:long-horizon-planning}
\small
\setlength{\tabcolsep}{4pt}
\renewcommand{\arraystretch}{1.12}
\begin{tabular*}{\linewidth}{@{\extracolsep{\fill}}llrrrr@{}}
\toprule
Task & Method & \multicolumn{4}{c}{Goal horizon (steps)} \\
\cmidrule(lr){3-6}
& & 25 & 50 & 75 & 100 \\
\midrule
\dataset{Push-T}
    & \method{}
    & $94.7\pm3.1$ & $60.7\pm3.1$ & $44.0\pm3.5$ & $43.3\pm4.6$ \\
& \methodname{LeWM (CEM)}
    & $68.7\pm3.1$ & $12.0\pm4.0$ & $3.3\pm2.3$ & $2.7\pm1.2$ \\
\midrule
\dataset{Cube}
    & \method{}
    & $100.0\pm0.0$ & $88.0\pm2.0$ & $80.7\pm2.3$ & $82.0\pm2.0$ \\
& \methodname{LeWM (CEM)}
    & $100.0\pm0.0$ & $0.0\pm0.0$ & $4.0\pm3.5$ & $4.0\pm3.5$ \\
\midrule
\dataset{Scene}
    & \method{}
    & $82.7\pm3.1$ & $73.3\pm8.1$ & $65.3\pm3.1$ & $56.0\pm4.0$ \\
& \methodname{LeWM (CEM)}
    & $44.7\pm13.0$ & $31.3\pm2.3$ & $37.3\pm5.8$ & $36.0\pm6.0$ \\
\midrule
\dataset{Puzzle}
    & \method{}
    & $74.0\pm12.5$ & $14.0\pm3.5$ & $4.0\pm3.5$ & $3.3\pm1.2$ \\
& \methodname{LeWM (CEM)}
    & $10.0\pm0.0$ & $4.7\pm2.3$ & $4.0\pm2.0$ & $0.7\pm1.2$ \\
\bottomrule
\end{tabular*}
\end{table}

\begin{table}[H]
\centering
\caption{\lead{Planning Speed across Four Visual-Goal Tasks.} Each plan is one
planner call. \methodname{LeWM (CEM)}~\citep{maes2026lewm} uses $300\times30$ CEM;
\method{} uses five gradient steps. Action dimensions are raw environment
action dimensions. Throughput is the
reciprocal of time per plan. The mean row averages per-task throughput, and its
ratio compares those two means.}
\label{tab:planning-speed}
\small
\setlength{\tabcolsep}{3.5pt}
\begin{tabular*}{\linewidth}{@{\extracolsep{\fill}}lcrrrrr@{}}
\toprule
& & \multicolumn{2}{c}{LeWM (CEM)} & \multicolumn{2}{c}{LeWAM} & \\
Task & Action dim. & ms/plan & plans/s & ms/plan & plans/s & Speedup \\
\midrule
\dataset{Scene}  & $5$ & $705$ & $1.42$ & $18.8$ & $53.19$ & $37.5\times$ \\
\dataset{Puzzle} & $5$ & $731$ & $1.37$ & $16.3$ & $61.35$ & $44.8\times$ \\
\dataset{Cube}   & $5$ & $771$ & $1.30$ & $21.1$ & $47.39$ & $36.5\times$ \\
\dataset{Push-T} & $2$ & $414$ & $2.42$ & $20.8$ & $48.08$ & $19.9\times$ \\
\midrule
Task mean & -- & -- & $1.62$ & -- & $52.50$ & $32.3\times$ \\
\bottomrule
\end{tabular*}
\end{table}

\begin{figure}[H]
  \centering
  \includegraphics[width=\textwidth]{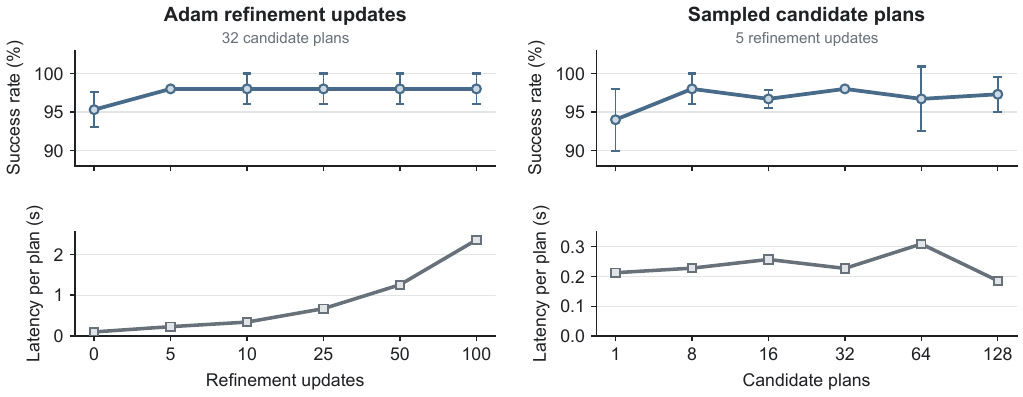}
  \caption{\lead{Planner Sensitivity on \dataset{Push-T}.} Goal-reaching success (mean and standard deviation across three seeds, 50 episodes each) and per-plan latency versus (left) Adam refinement updates with 32 candidate plans and (right) sampled candidate plans with five refinement updates.}
  \label{fig:pusht-grad-tradeoff}
\end{figure}

\lead{Planner Sensitivity.} In the \dataset{Push-T} sweep in
Figure~\ref{fig:pusht-grad-tradeoff}, success plateaus after five refinement
updates while latency continues to rise. Increasing the number of sampled
candidates beyond eight gives no consistent success gain in this sweep.

\subsection{Component and Visual-Encoder Ablations}
\label{app:component-ablations}

Table~\ref{tab:planning-component-ablation} records the SIGReg and action-head
comparisons in Figure~\ref{fig:model-ablations}(a,c). All variants use epoch-30
checkpoints. Table~\ref{tab:frozen-encoder-ablation} reports the separate encoder comparison
in Figure~\ref{fig:model-ablations}(b). Direct control executes the goal-conditioned
policy without search. Planning performs test-time action search through
the learned dynamics. The encoder panel reports equal-weight averages over
\dataset{Push-T}, \dataset{Cube}, \dataset{Scene}, and \dataset{Puzzle}.

\begin{table}[!htbp]
\centering
\caption{\lead{SIGReg and Action-Head Ablations.} Direct and planning success
rates (\%) using epoch-30 checkpoints for all variants. Gain is planning minus direct
success in percentage points. Standard deviations are reported where available;
none is inferred for gains.}
\label{tab:planning-component-ablation}
\small
\setlength{\tabcolsep}{4pt}
\renewcommand{\arraystretch}{1.1}
\begin{tabular*}{\linewidth}{@{\extracolsep{\fill}}llccc@{}}
\toprule
Environment & Variant  & Direct & Planning & Gain (pp) \\
\midrule
\multirow{3}{*}{\dataset{Push-T}}
 & \method{}  & $82.7\pm8.1$ & $97.3\pm3.1$ & $+14.7$ \\
 & w/o SIGReg  & $69.3\pm10.1$ & $72.0\pm3.5$ & $+2.7$ \\
 & MSE action head  & $90.0\pm4.0$ & $95.3\pm1.2$ & $+5.3$ \\
\midrule
\multirow{3}{*}{\dataset{Scene}}
 & \method{}  & $92.0\pm5.3$ & $97.3\pm1.2$ & $+5.3$ \\
 & w/o SIGReg  & $91.3\pm2.3$ & $93.3\pm3.1$ & $+2.0$ \\
 & MSE action head  & $89.3\pm4.2$ & $90.0\pm2.0$ & $+0.7$ \\
\midrule
\multirow{3}{*}{\dataset{Puzzle}}
 & \method{}  & $76.0\pm7.2$ & $80.7\pm11.0$ & $+4.7$ \\
 & w/o SIGReg  & $67.3\pm9.5$ & $75.3\pm9.0$ & $+8.0$ \\
 & MSE action head  & $76.7\pm6.1$ & $74.7\pm6.1$ & $-2.0$ \\
\midrule
\multirow{3}{*}{\dataset{Cube}}
 & \method{}  & $99.3\pm1.2$ & $100.0$ & $+0.7$ \\
 & w/o SIGReg  & $99.3\pm1.2$ & $99.3\pm1.2$ & $0.0$ \\
 & MSE action head  & $98.0\pm2.0$ & $99.3\pm1.2$ & $+1.3$ \\
\bottomrule
\end{tabular*}
\end{table}

\begin{table}[H]
\centering
\caption{\lead{Visual Encoders for Direct Control and Planning.} Success rate (\%)
with the same goal-conditioned evaluation protocol for all encoder variants.
Direct executes the policy without search; planning uses test-time action search.
Gain is planning minus direct success in percentage points. Averages are
equal-weight means over the four tasks; gains use unrounded means.}
\label{tab:frozen-encoder-ablation}
\footnotesize
\setlength{\tabcolsep}{2pt}
\renewcommand{\arraystretch}{1.1}
\begin{tabular*}{\linewidth}{@{\extracolsep{\fill}}llccccc@{}}
\toprule
Visual encoder & Mode & \dataset{Push-T} & \dataset{Cube} & \dataset{Scene} & \dataset{Puzzle} & Average \\
\midrule
\multirow{3}{*}{\method{} (end-to-end)}
 & Direct & $85.3\pm5.0$ & $100.0\pm0.0$ & $88.0\pm6.9$ & $73.3\pm5.8$ & $86.7$ \\
 & Planning & $98.7\pm1.2$ & $100.0\pm0.0$ & $93.3\pm3.1$ & $74.0\pm12.5$ & $91.5$ \\
 & Gain (pp) & $+13.4$ & $0.0$ & $+5.3$ & $+0.7$ & $+4.9$ \\
\midrule
\multirow{3}{*}{frozen DINOv2 CLS}
 & Direct & $72.7\pm7.6$ & $98.7\pm1.2$ & $82.0\pm4.0$ & $62.7\pm7.0$ & $79.0$ \\
 & Planning & $78.7\pm6.4$ & $99.3\pm1.2$ & $91.3\pm1.2$ & $61.3\pm13.0$ & $82.7$ \\
 & Gain (pp) & $+6.0$ & $+0.6$ & $+9.3$ & $-1.4$ & $+3.6$ \\
\midrule
\multirow{3}{*}{\shortstack[l]{frozen DINOv2\\patch mean}}
 & Direct & $66.0\pm12.5$ & $93.3\pm2.3$ & $84.0\pm2.0$ & $60.0\pm7.2$ & $75.8$ \\
 & Planning & $67.3\pm8.1$ & $98.7\pm1.2$ & $86.0\pm4.0$ & $58.7\pm7.6$ & $77.7$ \\
 & Gain (pp) & $+1.3$ & $+5.4$ & $+2.0$ & $-1.3$ & $+1.9$ \\
\bottomrule
\end{tabular*}
\end{table}

\FloatBarrier

\subsection{Probing Latent Representations}
\label{app:latent-probing}
Table~\ref{tab:physical-probing-appendix} extends the probing results presented on \dataset{Push-T} in Section~\ref{sec:physical-understanding} to \dataset{Cube, Scene, Puzzle}. All probes report $R^2$ from ridge regression on a $4000/1000$-frame train/test split. \emph{Object} is the mean $R^2$ over object ground-truth variables and \emph{Proprio} over proprioceptive variables. \emph{Action (IDM)} predicts the action chunk from $\vz_t$ (and from $[\vz_t,\vz_{t+K}]$, in parentheses where reported). \emph{Eff.\ rank} is the effective rank measure introduced in~\citep{royvetter2007effrank}.

\begin{table}[t]
\centering
\caption{\textbf{Effective Rank and Linear Probing.} Results on \dataset{Push-T}, \dataset{Cube}, \dataset{Scene}, and \dataset{Puzzle}, with the Object, Proprio, and Action columns reporting linear-probe $R^2$. Rows compare models with and without SIGReg; parenthesized values use both $\vz_t$ and $\vz_{t+K}$ for inverse-dynamics probing.}
\label{tab:physical-probing-appendix}
\small
\setlength{\tabcolsep}{6pt}
\renewcommand{\arraystretch}{1.10}
\begin{tabular}{@{}lllcccc@{}}
\toprule
Environment & Model & SIGReg & Eff. rank & Object $\uparrow$ & Proprio $\uparrow$ & Action (IDM) $\uparrow$ \\
\midrule
\multirow{5}{*}{\dataset{Push-T}} & \multirow{2}{*}{\method{}} & Yes & $42.90$ & $0.90$ & $\mathbf{0.86}$ & $\mathbf{0.40}\,(\mathbf{0.75})$ \\
 &  & No & $10.94$ & $0.88$ & $0.85$ & $0.38\,(0.73)$ \\
\cmidrule(l){2-7}
 & \method{} (MSE) & No & $6.96$ & $0.66$ & $\mathbf{0.86}$ & $0.36\,(0.74)$ \\
\cmidrule(l){2-7}
 & \multirow{2}{*}{\methodname{LeWM}} & Yes & $94.20$ & $\mathbf{0.95}$ & $0.80$ & $0.29\,(0.71)$ \\
 & & No & $16.23$ & $0.00$ & $0.00$ & $0.00\,(0.00)$ \\
\midrule
\multirow{3}{*}{\dataset{Cube}} & \multirow{2}{*}{\method{}} & Yes & $46.60$ & $\mathbf{0.99}$ & $0.67$ & $\mathbf{0.37}\,(0.66)$ \\
 &  & No & $8.00$ & $\mathbf{0.99}$ & $\mathbf{0.70}$ & $\mathbf{0.37}\,(\mathbf{0.68})$ \\
\cmidrule(l){2-7}
 & \methodname{LeWM} & Yes & $108.10$ & $\mathbf{0.99}$ & $0.58$ & $0.17\,(0.55)$ \\
\midrule
\multirow{3}{*}{\dataset{Scene}} & \multirow{2}{*}{\method{}} & Yes & $48.90$ & $\mathbf{0.48}$ & $\mathbf{0.78}$ & $0.39\,(\mathbf{0.74})$ \\
 &  & No & $12.50$ & $0.40$ & $0.77$ & $\mathbf{0.43}\,(\mathbf{0.74})$ \\
\cmidrule(l){2-7}
 & \methodname{LeWM} & Yes & $111.10$ & $0.40$ & $0.37$ & $0.07\,(0.35)$ \\
\midrule
\multirow{3}{*}{\dataset{Puzzle}} & \multirow{2}{*}{\method{}} & Yes & $46.50$ & $0.37$ & $\mathbf{0.70}$ & $\mathbf{0.33}\,(\mathbf{0.69})$ \\
 &  & No & $5.80$ & $0.22$ & $0.66$ & $0.31\,(0.68)$ \\
\cmidrule(l){2-7}
 & \methodname{LeWM} & Yes & $97.40$ & $\mathbf{0.60}$ & $0.42$ & $0.16\,(0.48)$ \\
\bottomrule
\end{tabular}
\end{table}

\label{tab:latent-probing-full}
\FloatBarrier

\subsection{Visual Reconstructions and Open-Loop Rollouts}
\label{app:simulation-rollouts}

\begin{figure}[p]
    \centering
    \includegraphics[width=\textwidth]{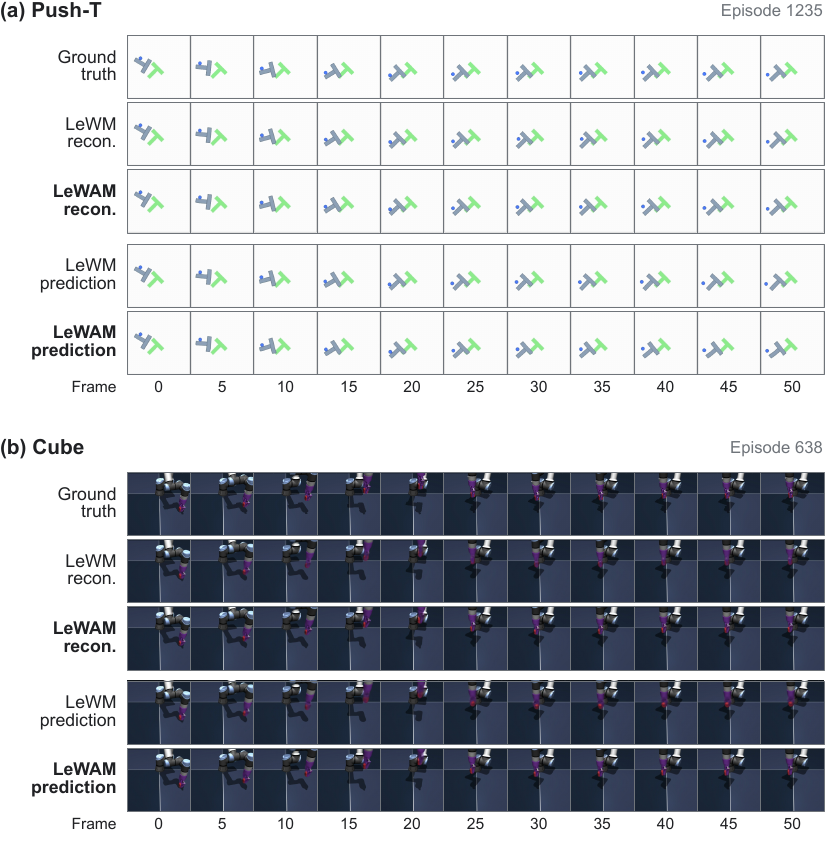}
    \caption{\lead{Aligned Visual Sequences on \dataset{Push-T} and \dataset{Cube}.}
    Each task shows recorded ground truth, framewise reconstructions for LeWM
    and LeWAM, and decoded open-loop predictions for both models. Columns use
    the same relative frame indices; predictions follow the recorded actions.}
    \label{fig:simulation-rollouts-pusht-cube}
\end{figure}

\begin{figure}[p]
    \centering
    \includegraphics[width=\textwidth]{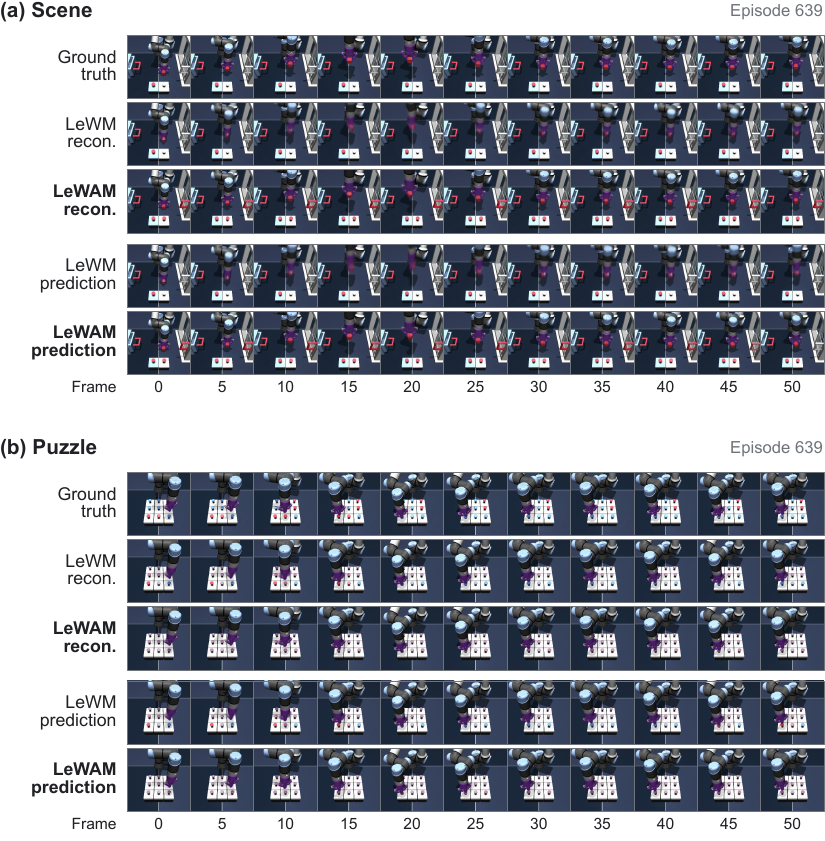}
    \caption{\lead{Aligned Visual Sequences on \dataset{Scene} and \dataset{Puzzle}.}
    Rows show recorded ground truth, LeWM and LeWAM framewise reconstructions,
    and their decoded open-loop predictions. Columns use the same relative frame
    indices and the same recorded action sequence for both prediction rows.}
    \label{fig:simulation-rollouts-scene-puzzle}
\end{figure}

Figures~\ref{fig:simulation-rollouts-pusht-cube}
and~\ref{fig:simulation-rollouts-scene-puzzle} align recorded frames with
pixel-decoded representations and latent predictions on
\dataset{Push-T}, \dataset{Cube}, \dataset{Scene}, and \dataset{Puzzle}.
Framewise reconstructions encode and decode each recorded observation independently. Open-loop predictions advance in five-step blocks from the initial observation context under fixed recorded actions. Each model receives the initial observed context and the recorded action sequence. We display the 51-frame clips at five-frame intervals and select the lowest-ID episode from the three recorded episodes available for each task.

The pixel decoders are trained post hoc on frozen world models using 150{,}000
sampled frames for 20 epochs; the displayed episodes are excluded from decoder
training. These diagnostic clips use 50-epoch LeWAM checkpoints on
\dataset{Cube}, \dataset{Scene}, and \dataset{Puzzle}, and an epoch-16 checkpoint
on the full \dataset{Push-T} dataset. LeWM uses our 15-epoch checkpoints from its
official implementation on the three 3D tasks and its released weights on
\dataset{Push-T}. Each model uses its own encoder, training schedule, and post-hoc pixel decoder.

On \dataset{Push-T}, both models reproduce pusher motion and block translation and rotation under the recorded actions (Figure~\ref{fig:simulation-rollouts-pusht-cube}). On \dataset{Cube}, \dataset{Scene}, and \dataset{Puzzle}, LeWAM preserves clearer robot-arm, gripper, and object details in its reconstructions and decoded predictions (Figures~\ref{fig:simulation-rollouts-pusht-cube} and~\ref{fig:simulation-rollouts-scene-puzzle}). These sequences visualize the physical-state information available to each model's learned dynamics. LeWAM has approximately 16.9M parameters, and \methodname{LeWM}~\citep{maes2026lewm} has 15M.

\lead{Planning with Learned Dynamics.} We remove \method{}'s action prediction head at evaluation time and use CEM~\citep{rubinstein1999cem} to plan with its action-conditioned latent dynamics. Table~\ref{tab:dynamics-only-cem} reports the resulting goal-reaching success rates alongside \methodname{LeWM}~\citep{maes2026lewm}.

\method{} achieves competitive success on \dataset{Push-T} and exceeds LeWM by $22.2$ percentage points on \dataset{Cube}. Together with the visual reconstructions and recorded-action predictions in Figure~\ref{fig:simulation-rollouts-pusht-cube}, these results show that its learned action-conditioned dynamics support planning, including contact-rich 3D manipulation.

\begin{table}[p]
    \centering
    \caption{\lead{CEM Planning with Learned Dynamics.} Goal-reaching success rates (\%), reported as mean $\pm$ standard deviation. LeWAM is evaluated with its action prediction head removed.}
    \label{tab:dynamics-only-cem}
    \small
    \setlength{\tabcolsep}{10pt}
    \renewcommand{\arraystretch}{1.10}
    \begin{tabular}{@{}lcc@{}}
        \toprule
        Model & \dataset{Push-T} & \dataset{Cube} \\
        \midrule
        \methodname{LeWM (CEM)} & $\mathbf{92.4\pm1.2}$ & $63.8\pm2.2$ \\
        \methodname{LeWAM (CEM)} & $84.0\pm1.3$ & $\mathbf{86.0\pm3.2}$ \\
        \bottomrule
    \end{tabular}
\end{table}

\FloatBarrier

\subsection{Multimodal Action Distributions}
\label{app:action-distributions}

\begin{figure}[H]
\centering
\begin{minipage}[t]{0.17\linewidth}
  \centering
  {\normalfont\scriptsize\sffamily (a) Data Trajectories\par}
  \vspace{3pt}
  \includegraphics[width=\linewidth]{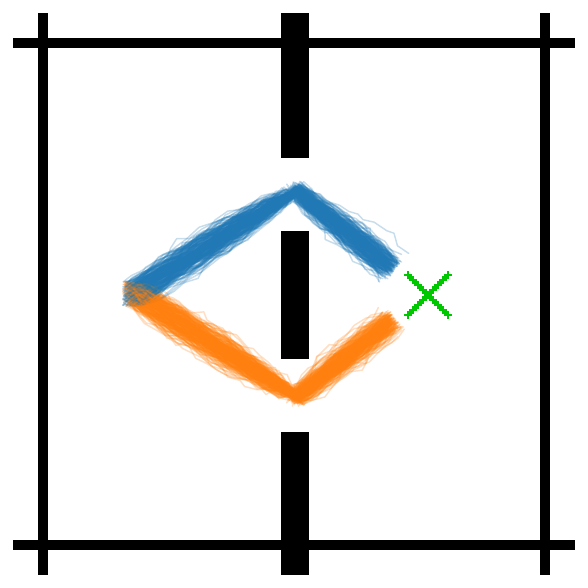}
\end{minipage}\hspace{0.02\linewidth}%
\begin{minipage}[t]{0.17\linewidth}
  \centering
  {\normalfont\scriptsize\sffamily (b) flow head\par}
  \vspace{3pt}
  \includegraphics[width=\linewidth]{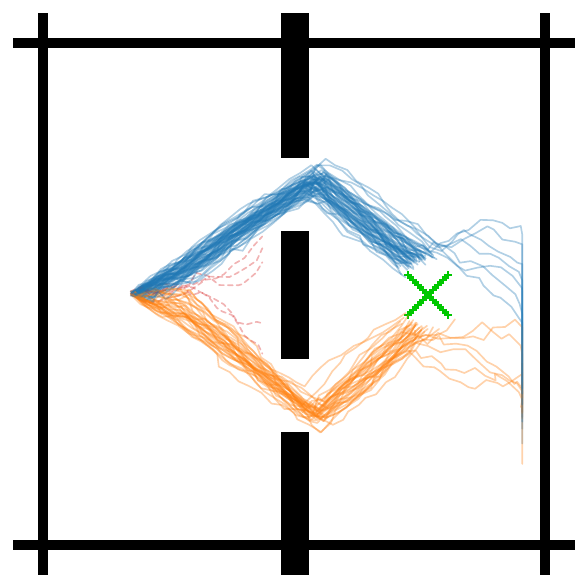}
\end{minipage}\hspace{0.02\linewidth}%
\begin{minipage}[t]{0.17\linewidth}
  \centering
  {\normalfont\scriptsize\sffamily (c) MSE head\par}
  \vspace{3pt}
  \includegraphics[width=\linewidth]{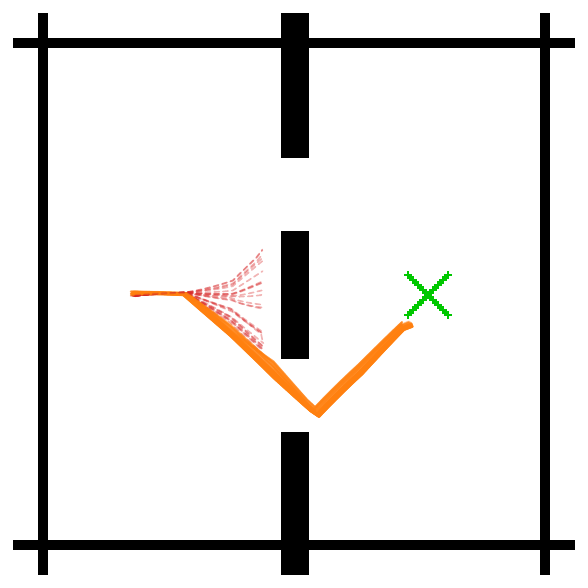}
\end{minipage}
\caption{\lead{Two-Door Demonstrations and Action-Head Rollouts.}
Expert demonstrations (a), executed trajectories from the flow-matching action
head (b), and executed trajectories from the deterministic MSE action head (c). Black
segments mark walls, and the green cross marks the goal. Blue and orange
indicate upper- and lower-door trajectories; dashed red paths end in collision.}
\label{fig:action-head-trajectories}
\end{figure}

To evaluate multi-route action distribution modeling, we collect expert demonstrations on a two-door navigation task with a fixed start and goal. Half of the expert trajectories pass through the upper door and half through the lower door, providing two valid routes between the same endpoints. We train \method{} using (1) our default flow-matching action head and (2) a deterministic MSE action head on the same data. During closed-loop policy rollouts (Figure~\ref{fig:action-head-trajectories}), the flow head follows both routes, whereas the deterministic MSE head
concentrates on the lower route and produces some wall collisions. The flow head preserves both route choices present in the demonstrations, illustrating how a stochastic action model can supply distinct proposals when a goal admits multiple routes.

\FloatBarrier

\subsection{Simulation Task-Completion Results}
\label{app:simulation-completion-results}
Table~\ref{tab:task-completion} reports the per-task results and deviations behind Figure~\ref{fig:model-ablations}(d).

\begin{table}[H]
\centering
\caption{Task completion. Mean success rates (\%) $\pm$ standard deviation across three evaluation seeds, with 50 episodes per seed. Bold marks the better result.}
\label{tab:task-completion}
\small
\setlength{\tabcolsep}{6pt}
\renewcommand{\arraystretch}{1.10}
\begin{tabular}{@{}lrrrr@{}}
\toprule
Method & DrawerCleanup & Transport & ToolHang & Average\\
\midrule
\methodname{Diffusion Policy}~\citep{chi2023diffusionpolicy}
    & $\mathbf{54.7\pm1.2}$ & $45.3\pm9.0$ & $\mathbf{84.7\pm3.1}$ & 61.6 \\
\method{} & $52.7\pm5.0$ & $\mathbf{61.3\pm9.9}$ & $80.0\pm3.5$ & $\mathbf{64.7}$ \\
\bottomrule
\end{tabular}
\vspace{-1em}
\end{table}
\FloatBarrier

\section{Physical-Robot Setup and Evaluation}
\label{app:real-robot}

This section documents the physical-robot experiments in Section~\ref{sec:behavior-cloning-robot-control}.

\subsection{Platform and Data Collection}
\label{app:robot-data}

Demonstrations are collected by joint-space teleoperation from a
\methodname{reBot Arm 102} leader to a \methodname{reBot B601-DM} follower through
\methodname{LeRobot}~\citep{cadene2026lerobot}. The follower uses impedance control
with gravity feedforward. One proficient operator collects successful demonstrations
without intentional exploration. Between episodes, both arms return to the calibration
zero pose before recording resumes.

A \methodname{ZED-Mini} records the scene at $672\times376$ pixels per eye, and a
wrist-mounted \methodname{Logitech C270} records at $640\times360$ pixels. The model
uses the left scene view and wrist view, both cropped to 256$\times$256 pixels; the right stereo view is recorded but unused.
At 30 Hz, a Raspberry Pi 5 stores the three video streams, seven-dimensional joint
targets as actions, and follower position, torque, and velocity as a 21-dimensional state.

\begin{table}[H]
\centering
\small
\caption{Datasets for the physical-robot tasks in Section~\ref{sec:behavior-cloning-robot-control}.}
\label{tab:robot-current-datasets}
\label{tab:robot-datasets}
\begin{tabular}{@{}lrrr>{\raggedright\arraybackslash}p{0.38\linewidth}@{}}
\toprule
Task & Episodes & Frames & Duration & Task Description \\
\midrule
Pick and Place T & 200 & 52,297 & 29 min & "Pick up the T block and place it onto the outline." \\
Push-T & 100 & 61,996 & 34 min & "Push the T block onto the outline." \\
Tape Stack & 100 & 23,726 & 13 min & "Stack the green roll of tape on the white roll of tape." \\
\bottomrule
\end{tabular}
\end{table}

\subsection{Model Implementation}
\label{app:robot-implementation}

The \method{} models in Section~\ref{sec:behavior-cloning-robot-control} use the state MSE and action flow
matching objectives from Section~\ref{sec:joint-predictor}. In the frozen
\methodname{V-JEPA 2.1} variant, attentive pooling and a BatchNorm MLP reduce each
camera frame to one summary token. Proprioceptive state and previous actions are
linearly projected from raw vectors to the model dimension. Learned queries initialize
the future-camera and action streams.

The real world \method{} variants take two video context frames and predict three future frames,
all sampled at 2.5~Hz.
They use model width 512 and depth eight, and train with AdamW
at batch size 256, using gradient accumulation when needed. Both variants use a
ViT-B visual backbone. The end-to-end model trains its visual encoder from scratch;
the frozen variant initializes its encoder from \methodname{V-JEPA 2.1}~\citep{murlabadia2026vjepa21}
and keeps it fixed. The action chunk length and execution interval are specified
in Appendix~\ref{app:bc-physical}.

For Table~\ref{tab:real-robot-results}, batch-1 GFLOPs include all inference steps
for one full action chunk: 100 DDPM steps for \methodname{Diffusion Policy}~\citep{chi2023diffusionpolicy},
ten Euler steps for $\pi_{0.5}$~\citep{physicalintelligence2025pi05}, and ten
Euler steps for each physical-robot \method{} variant.

\subsection{Physical-Robot Baseline Configurations}
\label{app:bc-physical}
For the comparison in Section~\ref{sec:behavior-cloning-robot-control}, both baselines use the
\methodname{LeRobot} implementations~\citep{cadene2026lerobot}, adapted to the
scene and wrist cameras and seven-dimensional joint-target actions. Each policy
returns 36 future actions and executes the first 18 at 30~Hz before requesting
a new chunk, matching the \method{} variants. The returned and executed chunks
span 1.2 and 0.6 seconds, respectively. The demonstrations and evaluation procedure
are described in Appendices~\ref{app:robot-data} and~\ref{app:robot-evaluation}, with 30 trials per task and method.

The \dataset{Pick and Place T} and \dataset{Push-T} training configurations resize both camera
streams to $256\times256$ and retain seven joint torques from the recorded
proprioceptive state. Actions and torques are normalized by mapping their
training-set first and 99th percentiles to $-1$ and $1$, respectively.
Training examples start every three recorded frames; the observation offsets
and action targets retain their 30~Hz timing. We hold out 10\% of episodes
using split seed zero. The configurations specify 20,000 optimizer updates
with effective batch size 256, a 500-update linear learning-rate warmup,
and cosine decay. Validation runs every 200 updates; the trainer saves both
the final checkpoint and the checkpoint with the lowest validation loss.

\baselineitem{MethodDiffusionPolicy}{\textbf{\methodname{Diffusion Policy}}~\citep{chi2023diffusionpolicy}.
The robot implementation uses a separate ImageNet-initialized ResNet-18 with
spatial softmax for each camera and a conditional one-dimensional U-Net with
channel widths $(336,672,1344)$. It conditions on two consecutive image and
torque observations. Images are randomly cropped to $243\times243$ during
training and center-cropped at evaluation. The internal diffusion horizon is
40 action positions, aligned from one step before the current observation;
the policy returns positions corresponding to the next 36 actions. This
40-step internal sequence satisfies the U-Net's temporal downsampling constraint.
Training uses noise-prediction MSE with a squared-cosine schedule, and inference
uses 100 DDPM steps. AdamW uses learning rate $10^{-4}$, betas $(0.95,0.999)$,
and weight decay $10^{-6}$. Exponential moving averaging with decay 0.999
supplies the weights used for validation and saved checkpoints.}

\baselineitem{PaperGray}{\textbf{$\pi_{0.5}$}~\citep{physicalintelligence2025pi05}.
The policy conditions on the current scene and wrist images together with
the task instruction and current torque vector. Its image frontend resizes
the inputs to $224\times224$. The wrapper discretizes the normalized torque
values into 256 bins and includes them in the language prompt.
The action expert predicts a 36-action chunk by flow matching, using ten
Euler steps at inference. The expert-only configuration freezes the vision
and language backbone while training the action expert and action/time
projections. Its task configurations use AdamW with learning rate
$2.5\times10^{-5}$ and weight decay $10^{-4}$, and gradient accumulation to
reach effective batch size 256.}

\subsection{Closed-Loop Evaluation}
\label{app:robot-evaluation}

 For Section~\ref{sec:behavior-cloning-robot-control}, the robot sends its observation window to a
remote inference machine, receives an action chunk, and executes it at 30 Hz under
the receding-horizon schedule. A human evaluator labels each episode as success or
failure, using the same criteria used during teleoperation. Each method is evaluated in 30 trials per task.

\subsection{Representation Diagnostics and Evaluation}
\label{app:robot-diagnostics}

For visualization, we fit a pixel decoder $d_\psi$ to frozen encoder outputs
$\vz_t=\enc(\vo_t)$. Learnable per-patch queries use cross-attention to map these
representations to pixel patches. Only the decoder parameters $\psi$ are trained,
using per-pixel MSE over training frames:
\begin{equation}
    \mathcal{L}_{\mathrm{dec}}(\psi)
    =\mathbb{E}_{\vo_t}\!\left[
    \operatorname{MSE}\bigl(d_\psi(\enc(\vo_t)),\vo_t\bigr)\right].
    \label{eq:robot-pixel-decoder}
\end{equation}
The encoder and latent dynamics remain frozen. Framewise reconstructions are
$d_\psi(\vz_t)$, and decoded rollout predictions are $d_\psi(\hat{\vz}_t)$.

\begin{figure}[htbp!]
    \centering
    {\small\bfseries End-to-end LeWAM\par}
    \includegraphics[width=\textwidth]{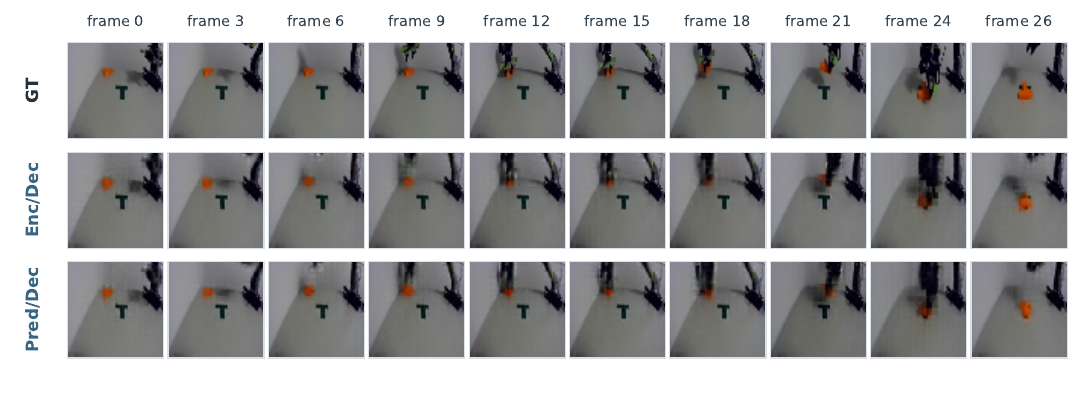}
    \vspace{2pt}
    {\small\bfseries LeWAM with frozen \methodname{V-JEPA 2.1}\par}
    \includegraphics[width=\textwidth]{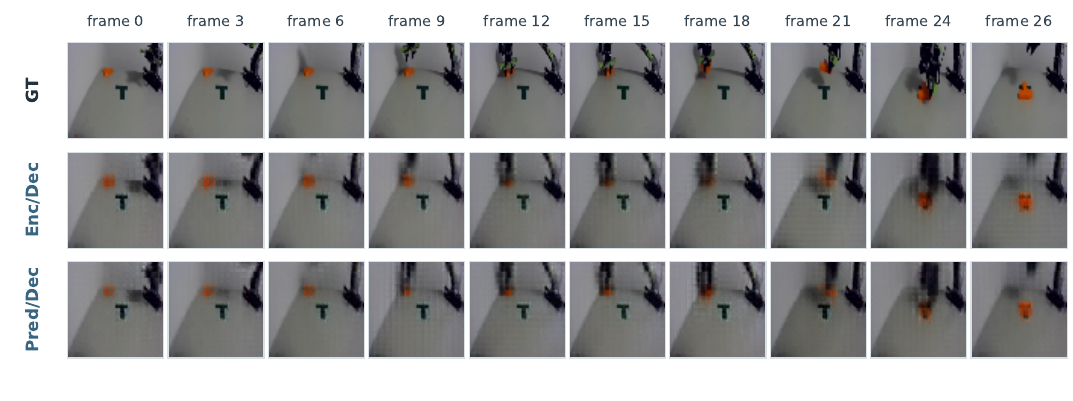}
    \caption{Matched visual reconstructions and recorded-action latent rollouts for the scene camera. Within each model, rows show the ground-truth (GT) video, independent frame reconstructions, and decoded autoregressive predictions.}
    \label{fig:robot-rollout-scene}
    \label{fig:robot-decode-panel}
\end{figure}

Figure~\ref{fig:robot-rollout-scene} compares both encoder variants on the same
held-out Pick and Place T episode from the scene camera. Each strip shows recorded
video, framewise reconstructions, and decoded autoregressive predictions. Latent
rollouts start from the initial observation context and advance under recorded
actions, without intermediate ground-truth frames or newly generated actions.
We display ten frames at indices 0, 3, 6, 9, 12, 15, 18, 21, 24, and 26, spanning
10.4 seconds at the 2.5~Hz model sampling rate.

\clearpage
\section{Discussion of the Role of Joint Prediction}
\label{app:joint-prediction}

The conditional-risk analysis describes how action supervision shapes representations; the planning discussion connects learned proposals to contact-dependent outcomes. These mechanisms complement Sections~\ref{sec:what-joint-buys} and~\ref{sec:control-planning}.

\Needspace{30\baselineskip}
\phantomsection
\label{app:system-simplicity}

Table~\ref{tab:world-action-factorizations} complements Section~\ref{sec:related} with a
comparison of training requirements and prediction capabilities.

\begin{table}[H]
\centering
\caption{\lead{Modeling and Control Capabilities.}
FDM denotes forward dynamics conditioned on clean actions; Policy denotes learned action
prediction. Entries describe reported training recipes or the goal-reaching configurations used here.}
\label{tab:world-action-factorizations}
\begingroup
\newcommand{\yes}{\ding{51}}
\newcommand{\no}{\ding{55}}
\setlength{\tabcolsep}{3pt}
\renewcommand{\arraystretch}{1.05}
\begin{tabular*}{\linewidth}{@{\extracolsep{\fill}}lcccc@{}}
\toprule
Method & FDM & Policy & \shortstack{Joint FDM\\and Policy} & \shortstack{End-to-end\\from scratch} \\
\midrule
\multicolumn{5}{@{}l}{\emph{World models}} \\
\methodname{DINO-WM}~\citep{zhou2024dinowm} & \yes & \no & \no & \no \\
\methodname{V-JEPA 2-AC}~\citep{assran2025vjepa2} & \yes & \no & \no & \no \\
\methodname{LeWM (CEM)}~\citep{maes2026lewm} & \yes & \no & \no & \yes \\
\midrule
\multicolumn{5}{@{}l}{\emph{World action models}} \\
\methodname{UVA}~\citep{li2025uva} & \yes & \yes & \no & \no \\
\methodname{UWM}~\citep{zhu2025uwm} & \yes & \yes & \no & \no \\
\methodname{DreamZero}~\citep{ye2026dreamzero} & \no & \yes & \no & \no \\
\methodname{Fast-WAM}~\citep{yuan2026fastwam} & \no & \yes & \no & \no \\
\methodname{Cosmos 3}~\citep{nvidia2026cosmos3} & \yes & \yes & \no & \no \\
\textbf{\method{} (ours)} & \yes & \yes & \yes & \yes \\
\midrule
\multicolumn{5}{@{}l}{\emph{Goal-conditioned policy}} \\
\methodname{LeWM (IDM)}~\citep{nguyen2026gcidm} & \yes & \yes & \no & \no \\
\methodname{GCBC}~\citep{ghosh2021gcsl} & \no & \yes & \no & \no \\
\bottomrule
\end{tabular*}
\par\vspace{3pt}
\begin{minipage}{\linewidth}
Joint requires clean-action FDM and policy losses on the same sample in one training pass.
From scratch includes visual encoding and prediction without pretrained or frozen components.
UWM uses a frozen pretrained variational autoencoder even when its robot backbone is trained from scratch.
\end{minipage}
\endgroup
\end{table}

\Needspace{12\baselineskip}
\subsection{Representation and Dynamics}
\label{app:joint-representation}

\begin{wrapfigure}{r}{0.40\linewidth}
\vspace{-16pt}
\centering
\includegraphics[width=\linewidth,trim=20bp 25bp 20bp 30bp,clip]{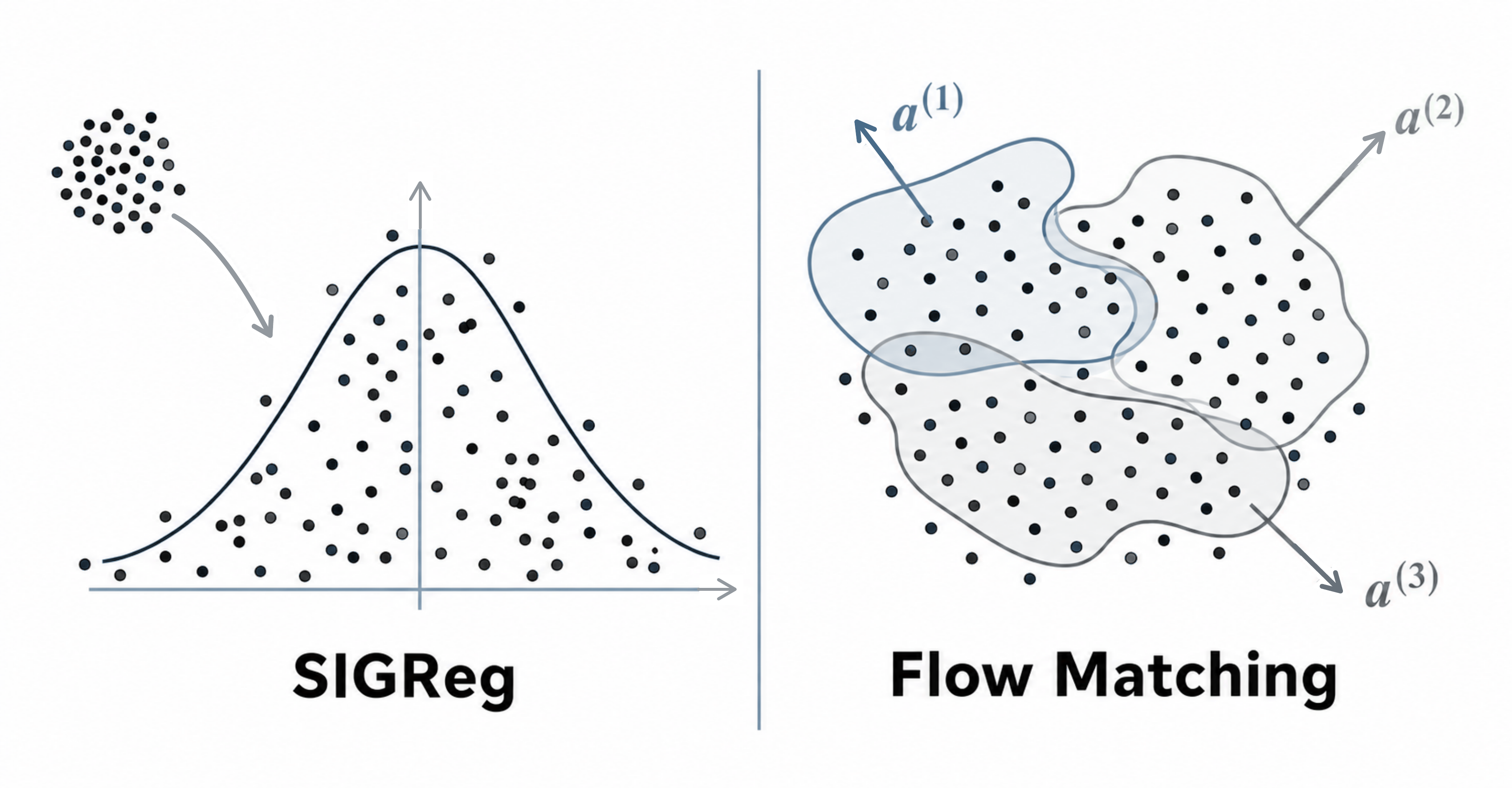}
\setlength{\abovecaptionskip}{3pt}
\vspace{-4pt}
\caption{Conceptual roles of SIGReg and action flow matching in shaping the latent space.}
\label{fig:anti-collapse-mechanisms}
\vspace{-6pt}
\end{wrapfigure}

\lead{Representation Shaping.} SIGReg~\citep{balestriero2025lejepa,maes2026lewm}
regularizes the latent marginal; action supervision encourages control-relevant
distinctions (Figure~\ref{fig:anti-collapse-mechanisms}). If visual history
changes the conditional action velocity, discarding it increases optimal action
regression risk. Latent prediction adds pressure to retain action-conditioned
consequences.
\finishwrapfigure

\Needspace{10\baselineskip}
\lead{Conditional Action Risk.} In Equation~\ref{eq:action-flow-path}, squared
flow matching makes the optimal velocity the conditional mean of
$\va_j-\vepsilon_j$. Let $\mathcal I_j$ include the noisy action prefix, flow
time, and other conditioning, but exclude visual history $\vz_{\leq t}$. With
$\mathcal R_{\mathrm{FM}}$ denoting optimal velocity MSE, the conditional
projection identity gives
\begin{equation}
    \begin{aligned}
    &\mathcal{R}_{\mathrm{FM}}(\mathrm{constant})-
      \mathcal{R}_{\mathrm{FM}}(\vz_{\leq t})\\
    &=\frac{1}{C}\sum_{j=t}^{t+C-1}\mathbb{E}\!\left[
      \operatorname{MSE}\!\left(
      \mathbb{E}[\va_j-\vepsilon_j\mid\mathcal I_j,\vz_{\leq t}],
      \mathbb{E}[\va_j-\vepsilon_j\mid\mathcal I_j]
      \right)
    \right]\geq 0.
    \end{aligned}
    \label{eq:flow-collapse-gap}
\end{equation}
When visual history changes the conditional velocity, a constant code incurs higher optimal risk. Action supervision therefore favors these distinctions, while SIGReg regularizes the global latent distribution.

\Needspace{24\baselineskip}
\subsection{Planning with Learned Action Proposals}
\label{app:joint-planning}

\begin{figure}[H]
\centering
\includegraphics[width=\linewidth]{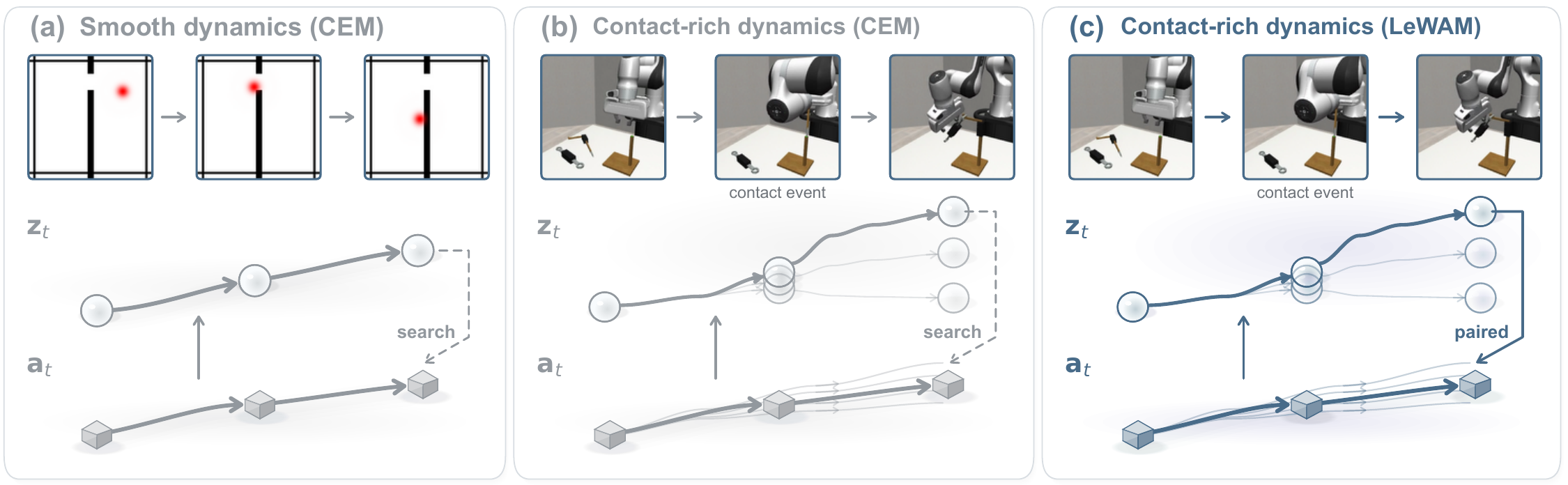}
\caption{\lead{Action Proposals and Predicted Consequences.}
\textbf{(a)} Smooth rollouts; \textbf{(b)} candidate rollouts branch at contact;
\textbf{(c)} LeWAM pairs learned action proposals with predicted outcomes.
Dashed links denote search; the solid link denotes joint prediction.
}
\label{fig:contact-geometry}
\end{figure}

\lead{Ill-Posed Action Solver in Contact-rich Events.} Prior JEPAs for control \eg, DINO-WM ~\citep{zhou2024dinowm} and \methodname{LeWM}~\citep{maes2026lewm}
learn action-conditioned latent dynamics without an action-prediction
objective and uses CEM~\citep{rubinstein1999cem} for downstream control. In smooth dynamics like 2D navigation, the action solver is \emph{well-posed}: a diverse representation can invert the action trajectories by argmin the cost of action-conditioned dynamics. However, a contact event can switch the active mode, so
nearby actions yield distinct outcomes~\citep{pang2023contactplanning}. If
sampled plans miss useful contact, an object-based goal cost provides little
ranking signal~\citep{yamada2024dcubed}. With a finite population, CEM ranks the contact branches it samples, and additional iterations incur more rollouts.

\lead{Behavior and Consequence.}  Joint prediction
additionally supplies a learned action distribution for planning proposals.
Learned proposals can improve contact-rich planning under
constrained budgets~\citep{brudermuller2025gpc}. Figure~\ref{fig:contact-geometry}
illustrates why the population-based search from a broad prior cannot work on contact-rich manipulation, and how LeWAM uses the learned proposals and predicted consequences to alleviate this problem. The
planning objective is defined in Section~\ref{sec:control-planning}, and the
refinement procedure is specified in Appendix~\ref{app:planners}. Though this problem in 3D contact-rich manipulation is well known and our solution is heuristic, we consider the rigorous verification of the claim as future work.

\clearpage
\section{Author Contribution Statement}
\label{app:author-contributions}
All coauthors contributed to project discussions, writing, and editing.

Minghao initiated the project and led its technical development. He proposed the
general architecture with action flow matching, MoT, the initial version of
gradient-based planning, and constructed the benchmarks. He also wrote the main
part of the paper and built the webpage.

Tavis designed the details of model architecture and MoT attention mechanism, optimized the planning algorithm to use batched gradients, and refined the methodology to the final version through extensive experimentation. He also finalized the released code. Tavis and Minghao
completed the simulation experiments reported in the paper.

Eryk designed and carried out the entire physical robot study. He built the hardware and teleoperation pipeline, adapted the model for real-robot deployment, collected all real-robot data, trained models for real robots, and ran the physical evaluations. He also wrote the main part of physical-robot experiments section and prepared its results.

Randall initiated and supervised the project and guided its overall direction. He advised on representation learning, experimental design, and result verification, contributed extensively to writing and refining the manuscript, and finalized the paper's title, narrative, and branding.

\end{document}